\documentclass[final,12pt]{clear2024} 

\usepackage{wrapfig}
\usepackage{booktabs}
\usepackage{multicol}
\usepackage[size=small]{caption}
\usepackage{multirow}

\newcommand{\paj}{\ensuremath{\textbf{PA}^\mathcal{G}_j}}

\DeclareMathOperator*{\argmin}{argmin}

\title[Robustness of Algorithms for Causal Structure Learning to Hyperparameter Choice]{Robustness of Algorithms for Causal Structure Learning to Hyperparameter Choice}
\usepackage{times}

\clearauthor{
 \Name{Damian Machlanski} \Email{d.machlanski@essex.ac.uk}\\
 \addr Department of Computer Science and Electronic Engineering, University of Essex
 \AND
 \Name{Spyridon Samothrakis} \Email{ssamot@essex.ac.uk}\\
 \addr Institute for Analytics and Data Science, University of Essex
 \AND
 \Name{Paul Clarke} \Email{pclarke@essex.ac.uk}\\
 \addr Institute for Social and Economic Research, University of Essex
}

\begin{document}

\maketitle

\begin{abstract}
Hyperparameters play a critical role in machine learning. Hyperparameter tuning can make the difference between state-of-the-art and poor prediction performance for any algorithm, but it is particularly challenging for structure learning due to its unsupervised nature. As a result, hyperparameter tuning is often neglected in favour of using the default values provided by a particular implementation of an algorithm. While there have been numerous studies on performance evaluation of causal discovery algorithms, how hyperparameters affect individual algorithms, as well as the choice of the best algorithm for a specific problem, has not been studied in depth before. This work addresses this gap by investigating the influence of hyperparameters on causal structure learning tasks. Specifically, we perform an empirical evaluation of hyperparameter selection for some seminal learning algorithms on datasets of varying levels of complexity. We find that, while the choice of algorithm remains crucial to obtaining state-of-the-art performance, hyperparameter selection in ensemble settings strongly influences the choice of algorithm, in that a poor choice of hyperparameters can lead to analysts using algorithms which do not give state-of-the-art performance for their data.
\end{abstract}


\section{Introduction}
Uncovering causal graphs is an immensely useful tool in data-driven decision-making as it helps understand the underlying data generating process. A large number of causal structure learning algorithms incorporate Machine Learning (ML) methods. These, in turn, heavily rely on hyperparameters (HPs) for accurate predictions \citep{bergstraAlgorithmsHyperParameterOptimization2011}. In addition, there has been growing evidence that correctly specified HPs can close the performance gap between State-of-the-Art (SotA) and other methods \citep{paineHyperparameterSelectionOffline2020,zhangImportanceHyperparameterOptimization2021,machlanskiHyperparameterTuningModel2023,tonshoffWhereDidGap2023}. \textit{Are hyperparameters as important in structure recovery?}

HP optimisation is extremely challenging in structure learning as the true graphs are inaccessible outside of simulated environments. This inability to reliably tune could be one of the reasons behind the struggle to apply some of the algorithms to real data problems \citep{kaiserUnsuitabilityNOTEARSCausal2021}, or why HPs are often completely neglected in this area. On the one hand, benchmarks and evaluation frameworks (e.g. \citet{raghuEvaluationCausalStructure2018a,tuNeuropathicPainDiagnosis2019a}) usually focus on finding a learning algorithm that works best under specific circumstances but without considering HPs as part of the problem. On the other hand, studies that address HP tuning (e.g. \citet{stroblAutomatedHyperparameterSelection2021,bizaOutofSampleTuningCausal2022}) consider individual algorithms but not the impact of tuning (or the lack of it) on selecting the best algorithm for the available data. 
Understanding how HPs affect algorithm choice, as well as individual methods, is clearly missing but can be a crucial next step towards more stable causal discovery in real data applications. To make matters worse, the evaluation metrics used for tuning can be imperfect and sometimes favour specific learning methods \citep{curthSearchInsightsNot2023}. This brings us to the core questions of this paper: \textit{Do different algorithms perform similarly given access to a hyperparameter oracle? How robust are they against misspecified hyperparameters?} 

\begin{figure}[t]
    \centering
    \includegraphics[width=\textwidth]{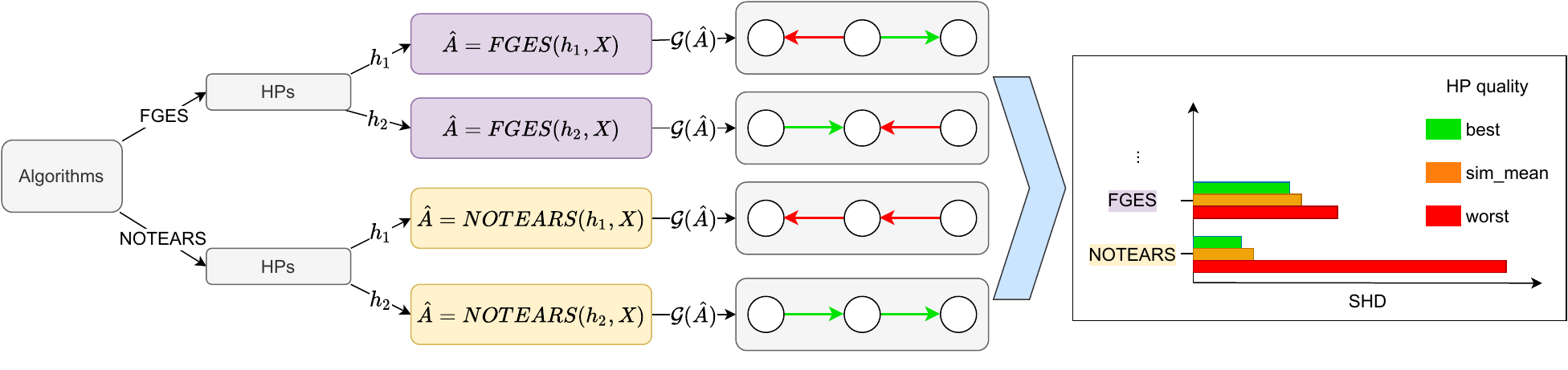}
    \caption{Summary of the main idea of the paper. We explore various structure learning algorithms and investigate how hyperparameters affect their performance. Notation: $h_1$ and $h_2$ are different hyperparameter values; $X$ denotes i.i.d. data provided to algorithms; $\hat{A}$ is recovered adjacency matrix while $\mathcal{G}(\hat{A})$ is a causal graph based on $\hat{A}$; SHD is structural Hamming distance (lower is better). Note how recovered graphs differ between different hyperparameters of the same algorithm (green edges are correct; red incorrect). \textit{sim\_mean} are hyperparameters that achieved the best \textbf{average} performance across all simulations.}
    \label{fig:summary}
\end{figure}

In this work, we set out to address these questions and investigate the impact HPs have on graph recovery performance of individual algorithms, as well as on the best algorithm choice (see Figure \ref{fig:summary}). We start by showing how a single HP plays a crucial role in the simplest graph problem (two variables). More extensive experiments strengthen this observation and confirm it as a more general phenomenon. The experimental setup involves many seminal structure learning algorithms tested against real and simulated datasets.

\noindent\textbf{Contributions.} a) Compare algorithms' performances and their winning percentages across hyperparameters, b) Compare algorithms' performances under well-specified and misspecified hyperparameters, and c) Compare algorithms' winning percentages under well-specified and misspecified hyperparameters.

\noindent\textbf{Related work.} This work connects with the existing literature mainly through the topics of performance evaluation and HP analysis. The performance of structure learning algorithms has been evaluated from a number of different perspectives, such as mixed data types \citep{raghuEvaluationCausalStructure2018a} or time series data \citep{assaad2022survey}. In an attempt to strengthen the evaluation, there have been efforts to develop testing environments that closely resemble real-life datasets. Some examples include simulators based on gene regulatory networks \citep{vandenbulckeSynTReNGeneratorSynthetic2006} or neuropathic pain pathology \citep{tuNeuropathicPainDiagnosis2019a}. \citet{grunbaumQuantitativeProbingValidating2023} take evaluation further by proposing to test algorithms on the parts of real-life datasets that are known a priori. Furthermore, to improve reproducibility, \citet{riosBenchpressScalableVersatile2022} developed a benchmarking platform that covers a wide range of learning methods and data scenarios. The importance of HPs and their impact on performance have been mostly studied in other areas outside of structure learning. In the offline Reinforcement Learning (RL) setting, \citet{paineHyperparameterSelectionOffline2020} reported, among other aspects, that robustness to HP choices is an important issue and that careful tuning can deliver close to optimal policies. \citet{zhangImportanceHyperparameterOptimization2021}, on the other hand, make a case for HP tuning in model-based RL. Similar observations have been reported in causal effect estimation \citep{machlanskiHyperparameterTuningModel2023}, graph learning \citep{tonshoffWhereDidGap2023}, code classification \citep{shen2020improving} and evolutionary algorithms \citep{eiben2011parameter}, where SotA performance levels are attributed to HPs alone. HPs have been also studied in the broader context of ``tunability'' \citep{probst2019tunability} and optimisation approaches \citep{yu2020hyper}. Some notable recent works even challenge our current understanding of how HPs interact with loss functions \citep{huang2023hyperparameter} and decision boundaries \citep{sohl2024boundary}. HPs in structure learning have mostly been discussed in the context of tuning. One common approach is to select HPs that result in stable structure predictions across random data samples \citep{liuStabilityApproachRegularization2010,sunConsistentSelectionTuning2013,stroblAutomatedHyperparameterSelection2021}. Another strand of work performs out-of-sample validation for tuning purposes based on predictive accuracy of models fitted in accordance with the recovered graph structure \citep{bizaOutofSampleTuningCausal2022}, or assigned scores developed specifically for structure recovery tuning \citep{chobthamTuningStructureLearning2023}. Metrics based on regression error have been also considered \citep{marxIdentifiabilityCauseEffect2019}, though in the context of two variables.

\noindent\textbf{Structure.} In Section \ref{sec:background} we briefly discuss the basics of structure learning. Section \ref{sec:hyperSL} demonstrates the importance of hyperparameters in structure learning via a bivariate example, further motivating more extensive numerical experiments presented in Section \ref{sec:exps}. Section \ref{sec:conclusion} concludes the paper and offers potential future work directions.

\section{Structure Learning}\label{sec:background}
We briefly describe the notation and data assumptions used throughout the paper as well as important details of structure learning methods necessary to read the technical parts of the document. For a more detailed review, see recent literature (e.g. \citet{eberhardtIntroduction2017,glymourReviewCausalDiscovery2019}).

\noindent\textbf{Graphs.} Let $\mathcal{G}=(V, \mathcal{E})$ be a graph with nodes/vertices $V=\{ 1,\ldots,p \}$ and edges $\mathcal{E} \subseteq V^2$. Edges are pairs of nodes $(j,k) \in \mathbf{V}$ where $(v,v) \notin \mathcal{E}$ to exclude self-cycles. Nodes $j, k$ are \textbf{adjacent} in $\mathcal{G}$ if either $(j,k) \in \mathcal{E}$ or $(k,j) \in \mathcal{E}$.  An edge is \textbf{undirected} if $(j,k) \in \mathcal{E}$ and $(k,j) \in \mathcal{E}$, whereas it is \textbf{directed} if only one pair appears in $\mathcal{E}$; if this pair is $(j,k)$ then $j$ is called a \textbf{child} of \textbf{parent} $k$. The set of parents of $j$ in $\mathcal{G}$ is denoted by \paj. We call $\mathcal{G}$ directed if all its edges are directed. A \textbf{mixed} graph consists of both directed and undirected edges. The \textbf{skeleton} of any directed or mixed graph $\mathcal{G}$ is an equivalent graph with all directed edges replaced by undirected ones. A \textbf{fully connected} graph $\mathcal{G}$ is one where all pairs of nodes are adjacent. A (directed) \textbf{path} is a sequence of nodes connected by (directed) edges. A \textbf{partially directed acyclic graph} (PDAG) is a mixed graph such that there is no pair $(j,k)$ such that there are directed paths from $j$ to $k$ and vice versa. Then, $\mathcal{G}$ is a \textbf{directed acyclic graph} (DAG) if it is a PDAG and is directed. Two graphs are \textit{Markov equivalent}, or belong to the same \textit{equivalence class}, when they involve the same sets of \textit{d-separations} \citep{pearlCausality2000}. A \textbf{completed PDAG} (CPDAG) can encode such a class of graphs, in which undirected edges mean that the graphs within the class may contain a directed edge in either direction; directed edges denote agreement in edge direction in subsumed graphs. 

\noindent\textbf{Assumptions}. Now consider a vector of random variables $\mathbf{X}=(X_1,\ldots,X_p)$ generated according to an unknown data generating process (DGP) leading to joint distribution $\mathcal{L}(\mathbf{X})$.  The node $j \in \mathbf{V}$ represents random variable $X_j$ and the edge between nodes $j$ and $k$ in $\mathcal{E}$ is directed if and only if $X_k$ is used in the DGP to generate $X_j$. We further assume that: a) there are no hidden confounders (\textit{sufficiency}); b) two variables are independent in $\mathcal{L}(\mathbf{X})$ if they are \textit{d-separated} in $\mathcal{G}$ (\textit{Markov condition}); and c) two variables are \textit{d-separated} in $\mathcal{G}$ if they are independent in $\mathcal{L}(\mathbf{X})$ (\textit{faithfulness}).

\noindent\textbf{Learning Methods.} The goal of causal structure learning is to infer (or identify) graph $\mathcal{G}$ given the distribution $\mathcal{L}(\mathbf{X})$. If it is possible to do this, we say $\mathcal{G}$ is \textbf{identifiable} from $\mathcal{L}$. Traditional methods, such as PC \citep{spirtesAlgorithmFastRecovery1991} or GES \citep{chickeringOptimalStructureIdentification2002}, were often built around assumptions a-c) above, which are in most cases not enough to identify a unique DAG solution, only the class of CPDAGs. If one, however, assumes the DGP is a Structural Causal Model (SCM) then identification is possible, for example, if the DAG is a linear SCM with additive non-Gaussian noise: $X_j = f_j(X_{\paj}) + \epsilon_j$ \citep{shimizuLinearNonGaussianAcyclic2006}. Subsequent research has extended this result to nonlinear SCMs with additive noise \citep{hoyerNonlinearCausalDiscovery2008}, linear models with Gaussian noise terms of equal variances \citep{petersIdentifiabilityGaussianStructural2014a}, and additive models of the form $X_j = \sum_{k \in \paj} f(X_k) + \epsilon_j$ \citep{buhlmannCAMCausalAdditive2014}. More recent approaches also involve neural network-based algorithms that specifically restate the learning task as a continuous optimisation problem \citep{zhengDAGsNOTEARS2018,zhengLearningSparseNonparametric2020}.

\section{Hyperparameters in Structure Learning}\label{sec:hyperSL}
\subsection{Bivariate Example}\label{sec:example}

\begin{wrapfigure}[20]{r}[0pt]{0.48\textwidth}
        \includegraphics[width=0.48\textwidth]{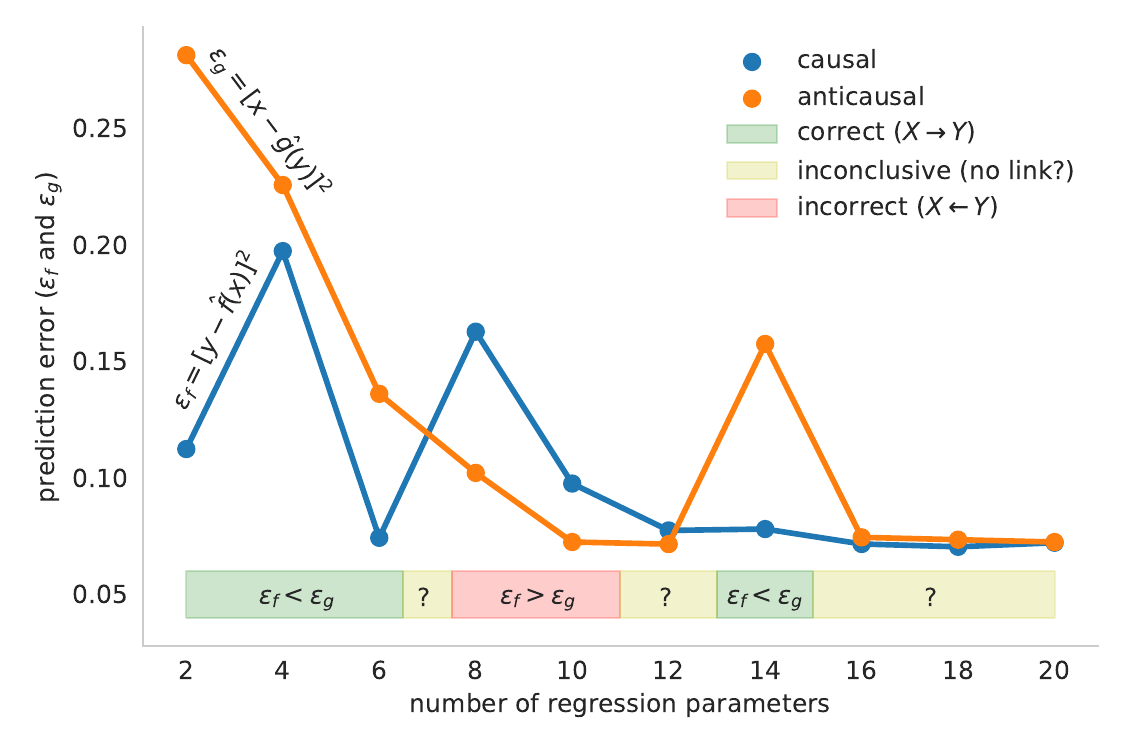}
    \caption{The number of allowed regression \textit{parameters}, a \textbf{hyperparameter}, clearly affects prediction error of the two models and can determine predicted causal direction (see decision colours at the bottom). The algorithm predicts $X \rightarrow Y$ if $\epsilon_f < \epsilon_g$, $X \leftarrow Y$ if $\epsilon_f > \epsilon_g$, and is inconclusive otherwise. Note that the true causal direction is unknown in practice.}
    \label{fig:example}
\end{wrapfigure}
An illustrative example, strongly inspired by \citet{marxIdentifiabilityCauseEffect2019}, is the classic \textit{cause-effect pairs} challenge \citep{guyon2019cause} that consists of two (synthetically generated here) variables $X$ and $Y$, with the goal of establishing the existence and direction of the causal link between them ($X \rightarrow Y$, $X \leftarrow Y$, no link) given only observed data. One possible solution is to fit two regressors $y=f(x)$ and $x=g(y)$, and predict the causal direction based on the lower prediction error of the two models ($\epsilon_f=[y - \hat{f}(x)]^2$ and $\epsilon_g=[x - \hat{g}(y)]^2$; no link if the errors are comparable). As shown in Figure \ref{fig:example}, changing the HP that controls the number of regression parameters can result in a different causal direction being predicted. This is precisely what constitutes the problem, since the true DGP and the correct HP value are unknown. \\
\noindent\textbf{Observation 1} \textit{Incorrect hyperparameters can cause prediction mistakes.} \\
\textbf{Observation 2} \textit{There might be more than one correct and incorrect hyperparameter choice.}

The problem grows in complexity as the number of graph nodes and edges increases. This is because each edge is a potentially different function to approximate that will require a different HP value (function complexity) to obtain the correct answer. In addition, many algorithms provide multiple HPs to tune, making even more room for further mistakes, effectively increasing the chance of HP misspecification. In fact, even the bivariate example can involve more HPs by, for instance, introducing a threshold such that the algorithm predicts `no link' if $| \epsilon_f - \epsilon_g | < threshold$. To this end, we make the following claim and set up the following key definition:

\noindent\textbf{Claim 1} \textit{The existence and direction of an edge in the predicted graph strongly depends on algorithm's hyperparameters.} \\
\textbf{Definition 1 (Hyperparameter Misspecification)} \textit{Mistakes in predicted graph structure arising from incorrect hyperparameters.}

Note that in practice HP misspecification may be not the only source of prediction mistakes as not all learning algorithms are guaranteed to converge to optimal solutions.

\subsection{General Form of the Problem}
Let $\mathbf{X} \in \mathbb{R}^{n \times p}$ represent i.i.d. data of $n$ observations and $p$ features. Furthermore, let $A \in \{ 0,1 \}^{p \times p}$ be a binary \textbf{adjacency matrix} of directed graph $\mathcal{G}(A)$ such that $a_{jk} = 1$ if $(j,k) \in \mathcal{E}$ and $a_{jk} = 0$ otherwise. A \textbf{weighted} adjacency matrix $W \in \mathbb{R}^{p \times p}$ is defined such that $A(W)_{jk} = 1$ if $w_{jk} \neq 0$ and zero otherwise, which results in a weighted graph $\mathcal{G}(W)$. Let us also define distance $d(A,B)$ between two adjacency matrices $A$ and $B$ such that $d(A,B)=0$ if and only if $\mathcal{G}(A) = \mathcal{G}(B)$ are the same graph. From now on, let us denote $A$ as the true adjacency matrix and $\hat{A}$ its estimate obtained by a programme $P$ from i.i.d. data $X$ using programme options $O$ so that
\begin{equation}
    \hat{A} = P(O, X),
\end{equation}
where $O$ generally involves the user specifying algorithm $S$ and hyperparameter values $H$. Therefore, given $K$ user-specified candidate programmes $P = \{ P_1,\ldots,P_K \}$ and 
\begin{equation}
    \hat{A}_k = P(S_k, H_k, X),
\end{equation}
where $S_k$ and $H_k$ are the algorithm and HP choices associated with program candidate $k$, then the best program is
\begin{equation}\label{eq:opt}
    k^* = \argmin_{k \in \{1,\ldots,K\}} d(A, \hat{A}_k),
\end{equation}
Note that $k^*$ is generally not identifiable unless $A$ is known. Furthermore, identification of $\mathcal{G}$ does not guarantee $\hat{A}_{k^*} = A$ as $\hat{A}_{k^*}$ depends on algorithms $S$ and their ability to identify $A$, as well as the choice of their HPs $H$.

More generally, when considering different algorithms $S$ and HPs $H$, Equation \ref{eq:opt} is the standard \textbf{model selection} problem, whereas if the choice of algorithms is fixed to a specific value, leaving HPs as the only variable, the task reduces to \textbf{hyperparameter tuning}. In practical terms, obtaining the distance $d(A,\hat{A})$ is not feasible, as the true matrix $A$ and its corresponding graph $\mathcal{G}(A)$ are inaccessible outside of simulated environments. As algorithms can vary substantially in design, the appropriate way to compare them requires the use of distance measures that incorporate the ground truth $A$. This renders model selection impractical in structure learning problems. Tuning HPs of a single algorithm might be feasible by comparing its relative scores across explored HP values.


\subsection{Common Hyperparameters}
Despite differences in algorithms, many of them share similar HPs. Commonly used ones are briefly described here.

\noindent\textbf{Significance level of independence tests.} Refers to the \textit{p-value} of independence tests and the desired level of certainty. Decreasing the value (increasing certainty) will usually result in fewer predicted edges. Often named as \textit{alpha} and incorporated by traditional and pairwise algorithms.

\noindent\textbf{Sparsity.} A penalty term that encourages sparser solutions. Higher values result in fewer predicted edges. Similar in mechanism to \textit{L1 regularisation} which discards less relevant features. Often employed by regression-based solutions, especially if they perform some form of feature selection.

\noindent\textbf{Model complexity.} A penalty that encourages simpler models to avoid overfitting (\textit{L2 regularisation}). As shown in the example in Section \ref{sec:example}, its influence on the final prediction is complicated. Usually applies to solutions that model the assumed form of SEMs.

\noindent\textbf{Post-prunning threshold.} Many SEM-based methods output weighted adjacency matrices $W$ that need to be converted to the binary form of $A$. This is usually done by applying a threshold below which all edges are set to zero. That is, $a_{ij} = 1$ if $w_{ij} > w\_thresh$; $0$ otherwise..

Note that \textit{alpha} and \textit{w\_threshold} are algorithm agnostic and can be transferred between methods, whereas the other two HPs may differ in value between algorithms.

\section{Experiments}\label{sec:exps}
Since our analysis in Section \ref{sec:hyperSL} is based merely on a simple and artificial example, our next step is to study the influence of HPs more rigorously in a more general setting. We devise a set of experiments consisting of diverse data sets of various size and difficulty (Section \ref{sec:data}), processed by a representative set of structure learning algorithms (Section \ref{sec:algs}). Different HP selection scenarios are also detailed (Section \ref{sec:hyper}).

The experimental framework is implemented through Benchpress \citep{riosBenchpressScalableVersatile2022}, a benchmarking platform to evaluate structure learning algorithms. Performances of all algorithms are collected from Benchpress, followed by some mild post-processing of the results to suit our analysis of HPs. The code and data are available via Appendix \ref{app:code}.

\subsection{Graphs and Data}\label{sec:data}
\subsubsection{Simulations}
We follow recent literature in structure learning when it comes to simulating different DGPs. The simulation procedure starts with generating a random DAG $\mathcal{G}$ with $p$ nodes and $d$ edges, built according to a random graph model. The resulting graph is binary, with $A \in \{0,1\}^{p \times p}$. Next, i.i.d. data $\mathrm{X} \in \mathbb{R}^{n \times p}$ are sampled from a simulated SEM of choice, with $n$ being the sample size. Each individual combination of settings is repeated for $10$ seeds and forms a single experiment. In our experiments, we explore $p \in \{10,20,50\}$, $d \in \{1p,4p\}$, and $n \in \{200,1000\}$. Included random graph models are Erd\"os-R\'enyi (ER) \citep{erdosRandomGraphs1959} and Barab\'asi-Albert \citep{barabasi1999emergence}, with the latter also known as scale-free (SF). We also explore $n=10,000$ but only for sparse ER graphs with $p=50$ nodes due to computational limitations. As for explored SEMs, we include the following:

\noindent \textbf{Linear Non-Gaussian} (\textbf{gumbel}). $X = X W^T + z \in \mathbb{R}^p$, with $W \in \mathbb{R}^{p \times p}$ as edge weights assigned independently from $U(\left [-2,-0.5 \right ] \cup \left [0.5,2 \right ])$ and based on $A$. Noise $z$ follows the Gumbel distribution $z \sim$ Gumbel($0, I_{p \times p}$).

\noindent \textbf{Nonlinear Gaussian} (\textbf{gp}). $X_j = f_j(X_{\paj}) + z_j$ for all $j \in \left [ p \right]$ in the topological order of $\mathcal{G}$. Noise $z_j$ follows Gaussian distribution $z_j \sim \mathcal{N}(0,1)$, $j=1,\ldots,p$. Where functions $f_j$ represent a draw from a Gaussian process with a unit bandwidth RBF kernel.

Note that both settings have been shown to be identifiable. That is, linear non-Gaussian additive models \citep{shimizuLinearNonGaussianAcyclic2006} and nonlinear additive models \citep{hoyerNonlinearCausalDiscovery2008}.

\subsubsection{Real Datasets}
We also tested structure learning algorithms against real or semi-real datasets. The most popular ones in the literature are \textit{protein signaling} and \textit{SynTReN}.

\noindent\textbf{Protein signaling} comes from \citet{sachsCausalProtein2005} which measures protein and phospholipid expression levels in human cells. The ground truth causal graph has been established and accepted by the experts in the field. We use the second dataset that is already logged and standardised and consists of $n=902$ observations, $p=11$ nodes and $d=17$ edges.

\noindent\textbf{SynTReN} is a generator of synthetic transcriptional regulatory networks and related gene expression data that simulate a real experiment \citep{vandenbulckeSynTReNGeneratorSynthetic2006}\footnote{http://bioinformatics.intec.ugent.be/kmarchal/SynTReN/index.html}. We use the same data as in \citet{lachapelleGradientBasedNeuralDAG2019}, which consist of $10$ random seeds, $n=500$ samples and $p=20$ nodes.

\subsection{Structure Learning Algorithms}\label{sec:algs}
We consider in our setup the following algorithms. PC \citep{spirtesAlgorithmFastRecovery1991}, FCI \citep{Spirtes1993}, FGES \citep{ramseyMillionVariablesMore2017}, LiNGAM \citep{shimizuLinearNonGaussianAcyclic2006}, ANM \citep{hoyerNonlinearCausalDiscovery2008}, CAM \citep{buhlmannCAMCausalAdditive2014}, NOTEARS \citep{zhengDAGsNOTEARS2018} and NOTEARS\_MLP \citep{zhengLearningSparseNonparametric2020}. Due to high computational demands, we only focus on well-established and seminal algorithms that, in our view, effectively represent different classes of solutions. More details about the algorithms and HPs involved can be found in Appendix \ref{app:algs}. Note the algorithms may have different termination criteria due to design differences, but they all produce (CP)DAGs.

\subsection{Evaluation}
We compare algorithms' performances across three commonly used metrics: \textit{structural Hamming distance} (SHD), \textit{false positives} (FPs), and \textit{false negatives} (FNs). All three accommodate for the fact that the ground truth is always a DAG but some of the incorporated algorithms output CPDAGs. The implementation of SHD we use counts not only the number of edge additions, removals and reversals, but also edge orientations, needed to turn the predicted graph into the true DAG. FPs and FNs count false edges and are calculated based on graph skeletons. For this purpose, predicted and true graphs are converted to skeletons to obtain the two metrics. This allows us to include methods that output CPDAGs even though the primary focus of this study is DAG recovery. See Appendix \ref{app:eval} for more detailed definitions of the metrics.

\subsection{Hyperparameters}\label{sec:hyper}
All incorporated learning algorithms have at least one HP. We collect performances of algorithms across all HP combinations (exhaustive grid search; see Appendix \ref{app:hyper}). Note that while many HPs are defined on continuous spaces, we search over a pre-defined set of points chosen to cover the space as completely as possible. To better understand the influence of HPs on structure recovery performance, we experiment with four different HP selection strategies described below.

\noindent\textbf{BEST.} To simulate the choice of the best HPs (as if we had access to the HP oracle), we pick HP values that achieved \textbf{the lowest} metric value in that particular data setting. Each data setting can have a different set of the best HPs.

\noindent\textbf{WORST.} Identified similarly to `best' except the criterion here is \textbf{the highest} metric value.

\noindent\textbf{DEFAULT.} Default HP values recommended by the authors of an algorithm. See Appendix \ref{app:algs}.

\noindent\textbf{SIM\_MEAN.} An alternative to `default'. We found a single set of HP values per algorithm that achieved \textbf{the lowest average} metric value across all simulations. These are \textit{simulation-derived} default values. See Appendix \ref{app:hyper} for identified values.

\subsection{Results}
Presented results employ the following naming convention with respect to the DGP: \textit{graph\_p} (number of nodes), \textit{graph\_d} (edge density), \textit{graph\_type} (graph models; ER or SF), \textit{data\_n} (sample size), \textit{data\_sem} (SEM type; gumbel or gp). Error bars are standard errors unless stated otherwise. Due to space limitation only the most important results are presented in the main content of the paper. The rest of the results, that do not change conclusions, are in Appendix \ref{app:results}.

\begin{figure}[htb]
    \centering
    \includegraphics[width=\textwidth]{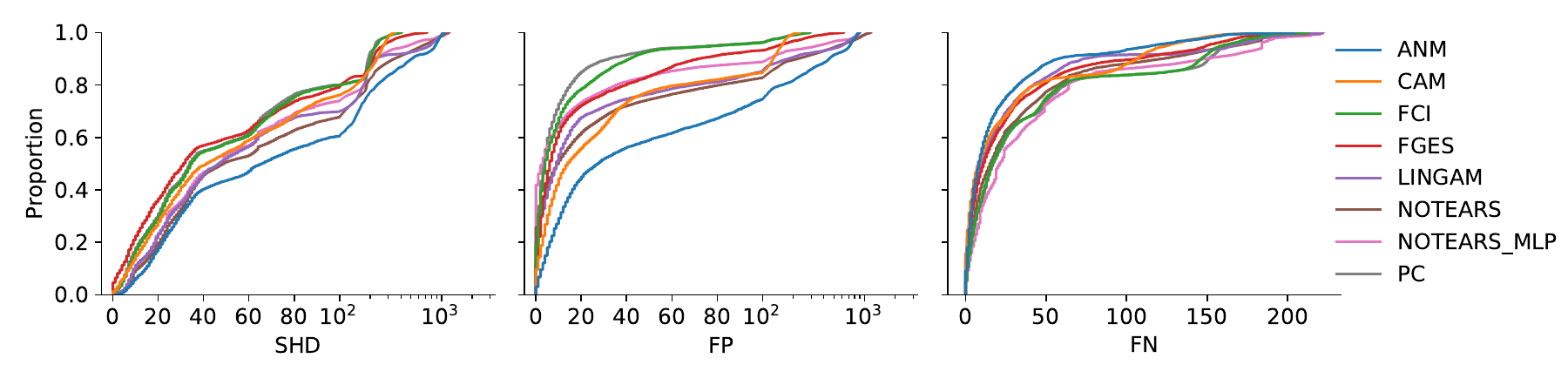}
    \caption{Proportions of performances (lower is better) across all HP values and simulations. Interpretation: algorithms perform similarly when averaged over all DGPs and HPs; no algorithm is the best in all conditions.}
    \label{fig:dist_combined}
\end{figure}

\noindent\textbf{Performance Distribution Across Hyperparameters.} As per Figure \ref{fig:dist_combined}, all algorithms perform similarly when averaged across all simulations and hyperparameters. This confirms that no single algorithm is the best in all conditions; they rather specialise in solving specific challenges that are ingrained in their design via assumptions. Some minor deviations from this general observation can be noticed when considering FPs (false positives) only (see FCI, PC and ANM), but they become negligible when considering the main metric (SHD).

\begin{figure}[!tb]
    \centering
    \includegraphics[width=\textwidth]{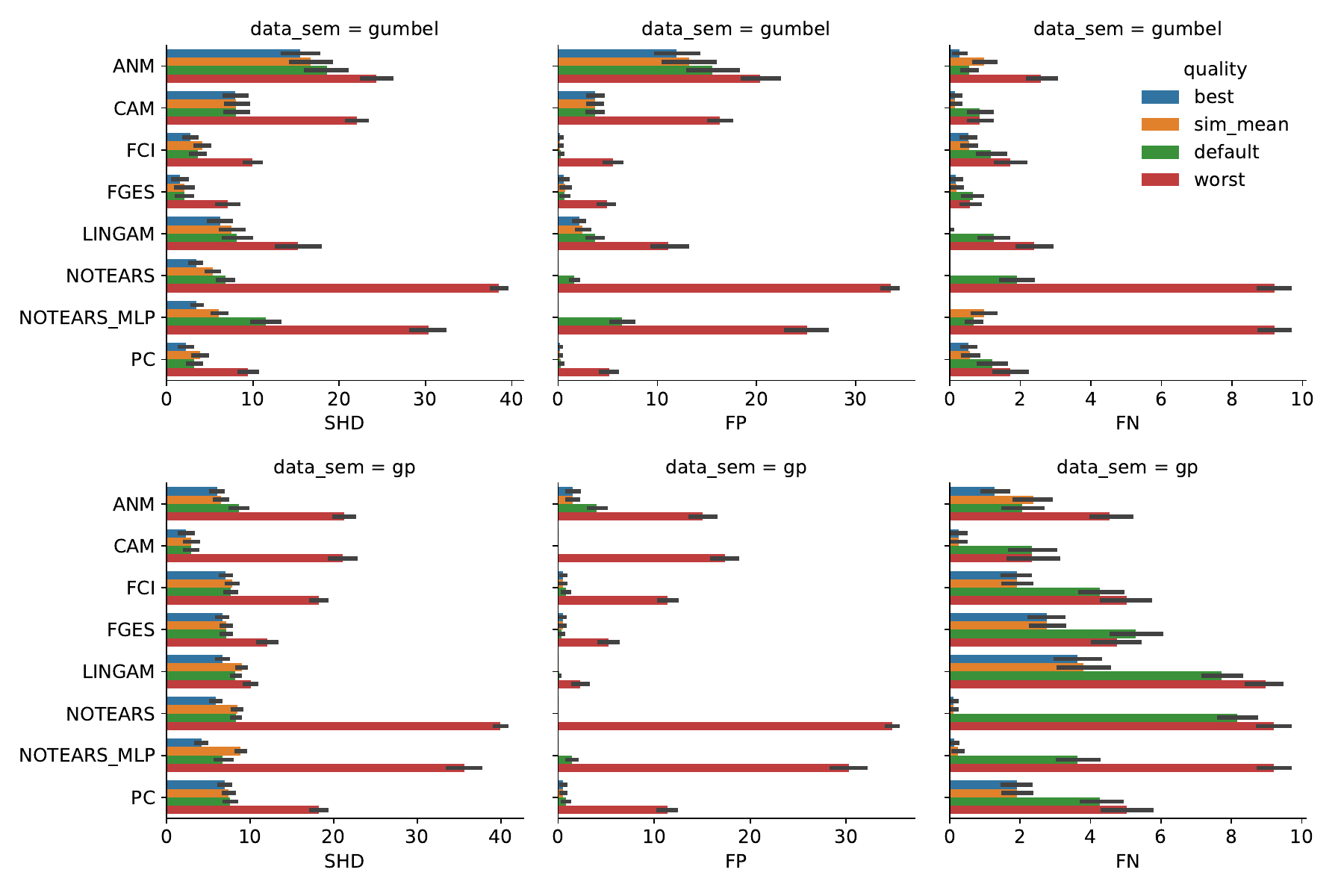}
    \caption{Performances (lower is better) for small sparse graphs ($p=10$, $d=1$) with linear (\textit{gumbel}) and nonlinear (\textit{gp}) SEMs depending on the \textbf{quality} of selected HPs (see the legend). Interpretation: a) fixed HPs (orange and green) perform similarly as the optimal ones (blue), b) algorithm selection is important even with optimal HPs (blue bars differ among algorithms), c) certain algorithms and setups are more risky (differences among red bars), and d) methods with optimised HPs provide only true edges (see blue bars in FPs).}
    \label{fig:hq}
\end{figure}

\noindent\textbf{Performance vs. Hyperparameter Quality.} As shown in Figure \ref{fig:hq}, achieved performance is clearly affected by both algorithm and HP choices. Even assuming access to HP oracle (blue colour), selecting different algorithms will have a significant impact on the result (blue bars differ among algorithms). This is because assumptions, either implicit or explicit, about the DGP play a critical role in controlling the performance of each method. It is worth noting that, in contrast to previous studies of HP choice for other causal methods, the absolute performance level achieved by the algorithms is low: no algorithm comes close to achieving $\mathrm{SHD}=0$. The complexity of the learning task is such that accurate structure discovery currently lies beyond the state of the art. In this context, fixed HPs (orange and green) seem to be a viable strategy as they are relatively close to the best cases (blue). This shows that HP values transfer well between different DGPs, which can be exploited in practice as a countermeasure to challenging HP optimisation. The differences between simulation-derived and paper-default values (orange and green respectively) are negligible in most cases with the exception of FNs (false negatives) where the former perform better. This indicates that the recommended defaults often prioritise sparsity, which reduces FPs, at the cost of increased FNs. The worst HPs (red), on the other hand, can result in performances substantially worse than the fixed ones. This shows the risks of HP misspecification, the degree of which clearly varies across algorithms (different robustness), which could be due to again different data assumptions in algorithms or varied degrees of freedom via algorithm's HPs. Note also how the majority of mistakes under misspecified HPs (red) is due to FPs (red bars in FPs larger than in FNs). However, once HPs are optimised (blue), FPs are mostly eliminated and the remaining mistakes are now due to FNs (blue bars larger in FNs than FPs). This has critical implications for practice: the predicted edges can be trusted (FP small) but the absence of an edge cannot (FN large). Remaining FNs could be also a sign of too aggressive sparsity/prunning or simply failed identification, with the latter being an indicator for future algorithmic improvements.

\begin{figure}[!tb]
    \centering
    \includegraphics[width=\textwidth]{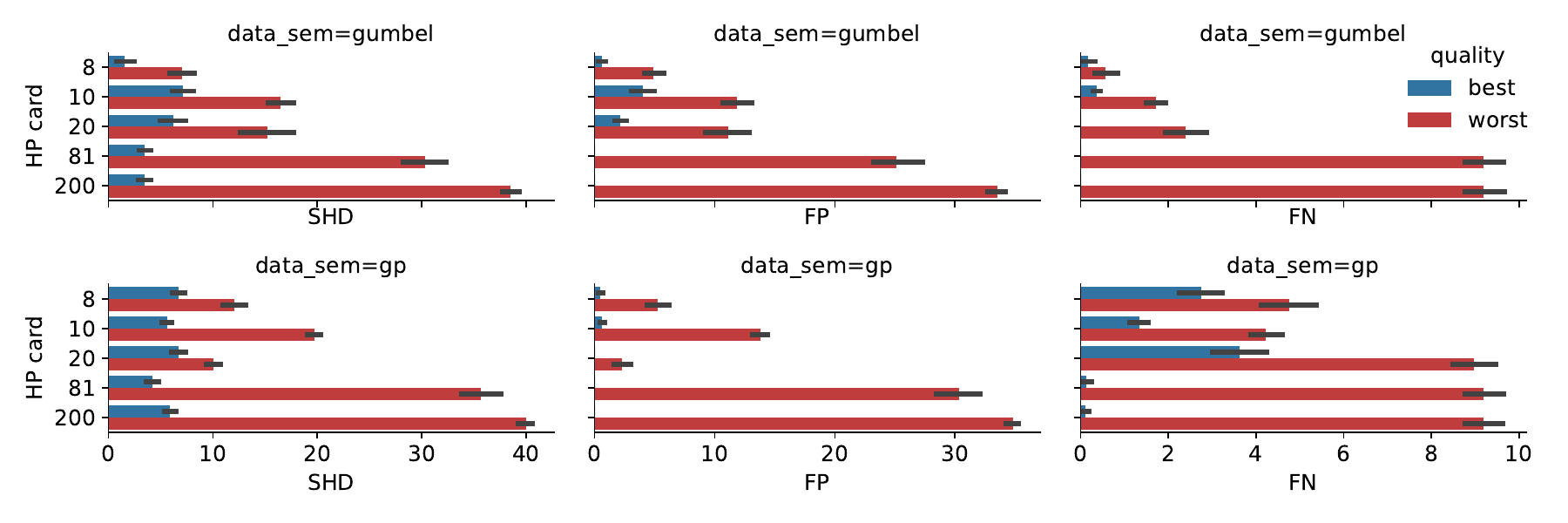}
    \caption{The influence of explored HP cardinalities (\textit{HP card}) across algorithms on performance (lower is better), depending on the \textbf{quality} of selected HPs (see legend). DGP: small sparse graphs ($p=10$, $d=1$) with linear (\textit{gumbel}) and nonlinear (\textit{gp}) SEMs. Note how higher cardinalities lead to small SHD gains under optimal HPs (blue), but at the cost of much worse performances under misspecified HPs (red). Warning: different cardinalities may involve different algorithms (possible coincidental trends).}
    \label{fig:card_all}
\end{figure}

\noindent\textbf{Performance vs. Cardinality of Hyperparameters.} A speciously interesting comparison that emerges from our study is the relationship between the cardinality of the HPs we explore and algorithm performance. Our study was not set up to explore this issue so we now clarify what can and cannot be concluded about this relationship from our study. As presented in Figure \ref{fig:card_all}, higher cardinalities ($81$ and $200$) lead to only small performance improvements under optimised HPs (blue bars), but they involve significantly higher risks of poor performances if HPs are misspecified (red). This shows that larger HP search spaces should be explored with caution as they provide little gain for a much higher risk of poor performances. Note, however, that different cardinalities presented here may involve different algorithms, making room for coincidental trends. For example, it is unclear given the data if robustness to misspecified HPs is due to HP cardinalities or algorithms as the two highest cardinalities were explored exclusively with (potentially volatile) neural networks.

\begin{table}[b]
\caption{Top algorithm choices based on the highest winning percentages, grouped by DGP properties (graph\_p, data\_sem, graph\_d) and quality of selected HPs (best, sim\_mean, worst). If there is no clear winner, multiple top performers are reported. Notice how the winners change across DGPs (columns), but also across HPs (rows), showing algorithm selection is nuanced. N\_MLP denotes NOTEARS\_MLP.}
\label{tab:ms_h}
\resizebox{\textwidth}{!}{%
\begin{tabular}{@{}rllll|llll|llll@{}}
\toprule
graph\_p & \multicolumn{4}{c|}{10} & \multicolumn{4}{c|}{20} & \multicolumn{4}{c}{50} \\
data\_sem & \multicolumn{2}{c}{gumbel} & \multicolumn{2}{c|}{gp} & \multicolumn{2}{c}{gumbel} & \multicolumn{2}{c|}{gp} & \multicolumn{2}{c}{gumbel} & \multicolumn{2}{c}{gp} \\
graph\_d & \multicolumn{1}{c}{1} & \multicolumn{1}{c}{4} & \multicolumn{1}{c}{1} & \multicolumn{1}{c|}{4} & \multicolumn{1}{c}{1} & \multicolumn{1}{c}{4} & \multicolumn{1}{c}{1} & \multicolumn{1}{c|}{4} & \multicolumn{1}{c}{1} & \multicolumn{1}{c}{4} & \multicolumn{1}{c}{1} & \multicolumn{1}{c}{4} \\ \midrule
\begin{tabular}[c]{@{}r@{}}HP\\ best\end{tabular} & \begin{tabular}[c]{@{}l@{}}FGES\\ FCI\end{tabular} & FGES & CAM & CAM & FGES & \begin{tabular}[c]{@{}l@{}}PC\\ N\_MLP\end{tabular} & CAM & CAM & FGES & \begin{tabular}[c]{@{}l@{}}PC\\ N\_MLP\end{tabular} & CAM & CAM \\ \midrule
\begin{tabular}[c]{@{}r@{}}HP\\ sim\_mean\end{tabular} & \begin{tabular}[c]{@{}l@{}}FGES\\ PC\end{tabular} & \begin{tabular}[c]{@{}l@{}}FGES\\ CAM\\ ANM\end{tabular} & CAM & CAM & FGES & PC & CAM & CAM & FGES & PC & CAM & CAM \\ \midrule
\begin{tabular}[c]{@{}r@{}}HP\\ worst\end{tabular} & FGES & \begin{tabular}[c]{@{}l@{}}FGES\\ PC\\ CAM\end{tabular} & \begin{tabular}[c]{@{}l@{}}LiNGAM\\ FGES\end{tabular} & CAM & FGES & PC & LiNGAM & \begin{tabular}[c]{@{}l@{}}CAM\\ LiNGAM\end{tabular} & FGES & PC & LiNGAM & LiNGAM \\ \bottomrule
\end{tabular}%
}
\end{table}

\noindent\textbf{Winning Algorithms vs. Hyperparameter Quality.} Previous analysis revealed that the best algorithm choice may depend on specific DGP properties as well as HP choices. To make it clearer, we collect winning algorithms across different DGP properties and HP selection strategies. An algorithm with the lowest SHD in a given setting wins. Table \ref{tab:ms_h} presents top performers across different settings based on accumulated winning percentages. The results confirm that no single algorithm wins in all settings. Specific DGP properties may favour certain methods. When looking at how top performers change depending on different HP choices, it is clear that the best algorithm selection depends not only on DGP properties, but also on the type of available HPs. For instance, to minimise the risk of poor performances, one can select algorithms from the `HP worst' category.

\begin{figure}[htb]
    \centering
    \includegraphics[width=0.9\textwidth]{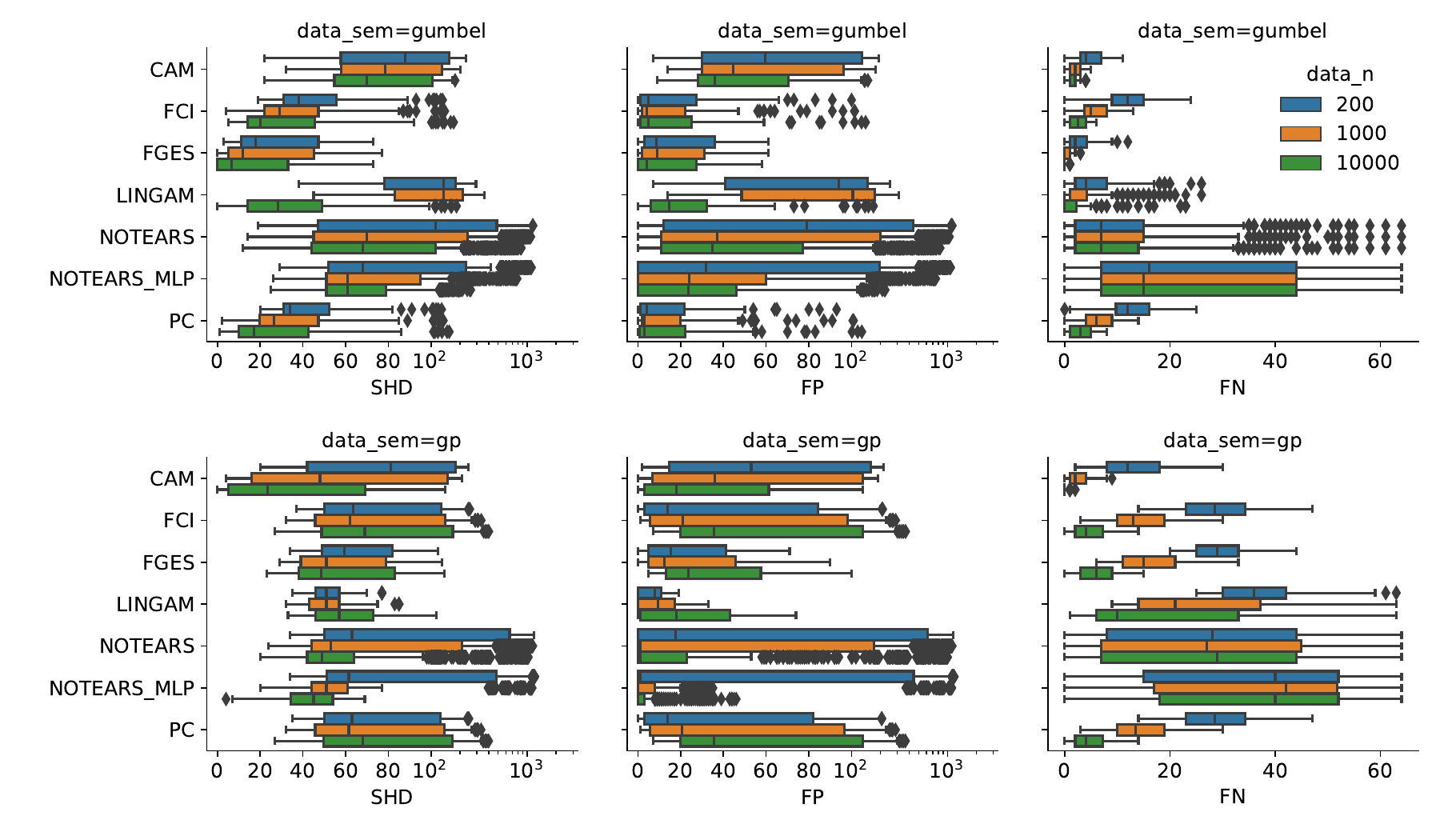}
    \caption{The influence of sample size (\textit{data\_n}; colours) on performances (lower is better) across all HPs. DGP: sparse ($d=1$) large ($p=50$) ER graphs with linear (\textit{gumbel}) and nonlinear (\textit{gp}) SEMs. ANM was excluded due to long execution time against $10,000$ samples. Note how increased sample size not only improves performances under the best and worst HPs (min and max decrease), but also the proportion of good performances (means decrease as well), suggesting increased sample size helps with HP misspecification.}
    \label{fig:dist_10k}
\end{figure}

\noindent\textbf{Performance vs. Sample Size.} Figure \ref{fig:dist_10k} confirms that increased sample size generally helps, even with relatively large graphs ($p=50$), though some algorithms need more data to notice major benefits (see LiNGAM in gumbel and NOTEARS\_MLP in gp). Positive effects can be noticed with respect to improved best HP cases (min values) and an increased proportion of good performances (mean values). This case also confirms that relatively large and sparse graphs can be recovered with high accuracy given the right HPs (min values of some green boxes are close to $0$).

\begin{figure}[t]
    \centering
    \subfigure[SynTReN dataset. Numbers are averaged across data seeds.]{\includegraphics[width=\textwidth]{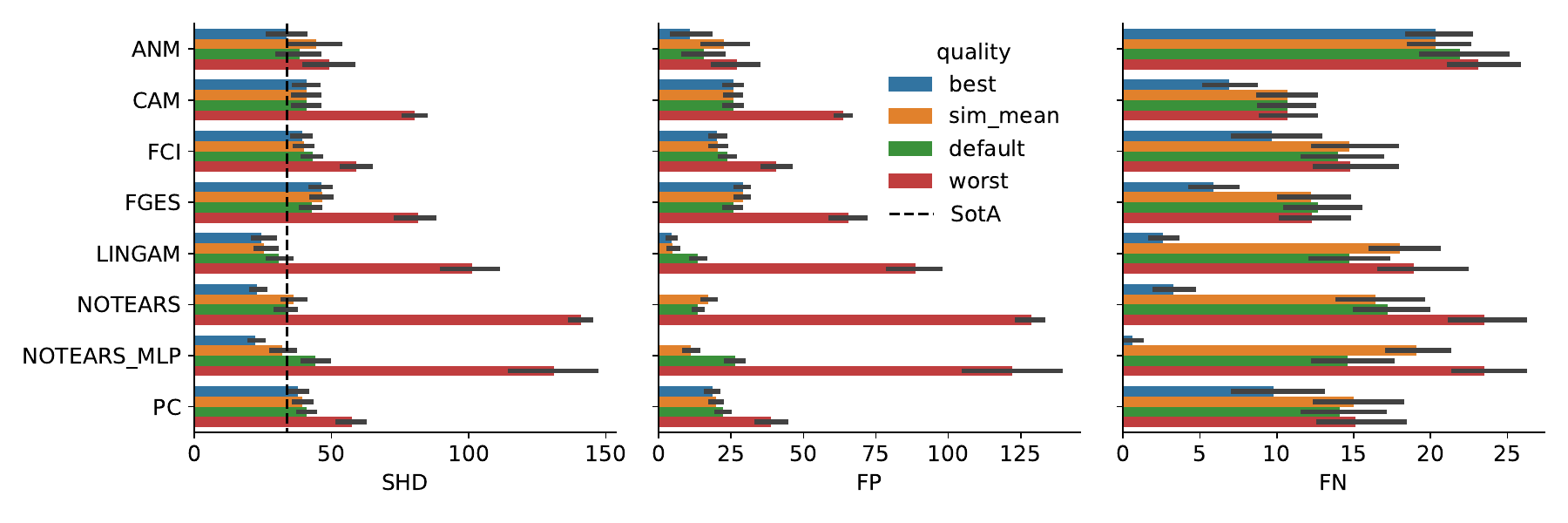}\label{fig:syntren}}
    \subfigure[Sachs dataset]{\includegraphics[width=\textwidth]{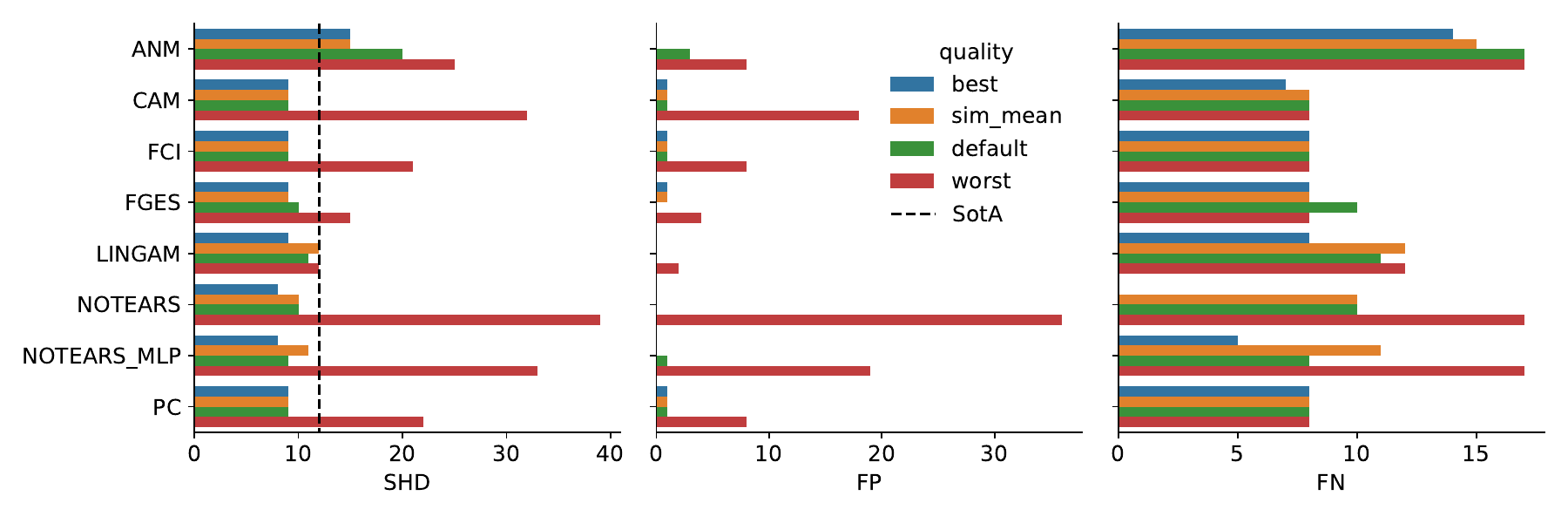}\label{fig:sachs}}
    \caption{Performances (lower is better) against SynTReN (top) and Sachs (bottom) datasets depending on the \textbf{quality} of selected HPs (colours). Notice that: a) HP values derived from simulations perform well here (orange), and b) the quality of HPs and algorithm choice are both important for beating SotA.}
    \label{fig:real_data}
\end{figure}

\noindent\textbf{Semi-Synthetic and Real Data.} We put our simulation-derived findings to a test by performing structure recovery on SynTReN and Sachs datasets (Figure \ref{fig:real_data}). SHD numbers are compared to SotA performances retrieved from \citet{lachapelleGradientBasedNeuralDAG2019}, which are $33.7 \pm 3.7$ and $12$ SHD for SynTReN and Sachs respectively. Both cases generally confirm that fixed HPs (\textit{sim\_mean} and \textit{default}) can work almost as well as the best HPs, and that even the best HPs may not be enough to reach the best possible performance as some algorithms perform better than others under those conditions. It is also clear from both cases that HPs play an important role and, in fact, can decide whether an algorithm reaches or beats SotA. For instance, against SynTReN, both NOTEARS methods and LiNGAM seem to be good options under the best and fixed HPs. But under the worst HPs, NOTEARS methods can be extremely inaccurate, making ANM the safest choice in this case. In the Sachs dataset, this is no longer the case with ANM, showing that the best algorithm pick indeed strongly depends on DGP properties. All algorithms except ANM can, in fact, beat SotA on Sachs data. However, when it comes to robustness to HP misspecification and safety of use, NOTEARS methods appear to be the most risky, with LiNGAM being extraordinarily robust as it beats SotA even with its worst HPs.

\section{Conclusion}\label{sec:conclusion}
In this work, we have successfully shown that HPs play an important role in causal structure learning. However, the way HPs influence the methods is somewhat different than recent results from the ML literature. More specifically, \citet{machlanskiHyperparameterTuningModel2023,tonshoffWhereDidGap2023} found that many learners can reach SotA performance levels with the right HPs, reducing the importance of model selection. But in this study, we observe that algorithms still differ significantly in performance even with access to the HP oracle. However, reliable tuning is not always available in structure learning, leading to HP misspecification and prediction errors. This is where HPs become important as we showed that different learning algorithms vary in robustness to HP misspecification, and that strong performance under the right HPs does not necessarily translate to misspecification robustness. As a consequence, an algorithm that is the best pick under correct HPs, might be a suboptimal choice when its HPs are misspecified; another algorithm with better misspecification robustness might be safer to use, especially in those cases where minimising the consequences of potential misspecification is a priority. Thus, overall, the best algorithm choice may depend not only on the properties of the data generating process, but also on the quality of selected HP values. In terms of secondary findings, default HPs seem to perform surprisingly well in many cases, and hence may constitute a viable alternative to tuning. Moreover, in the case of sparse graphs, predictions with optimised HPs seem to include only true edges (no FPs, some FNs), which has critical implications for practice. Another interesting observation is that relatively large sparse graphs ($50$ nodes) can be recovered with high accuracy, subject to large sample size and the right HPs. It is also important to acknowledge the possibility that robustness to misspecified HPs might be impacted by the cardinality of explored HPs as larger search spaces may increase the probability of poor performances. Our results show this is indeed possible, but crucially the chances of good performances also slightly rise in this case. This suggests that higher HP cardinalities can be advantageous and disadvantageous, which motivates including the worst performances in this analysis. Furthermore, while we agree that the probability of getting the worst performances is low in practice, it is undeniably greater than zero and hence worth considering.\\
\noindent\textbf{Recommendations.} As for practical advice, we stress the fact that there are no universal answers in structure learning; there are many forces at play (DGP properties, algorithms, HPs) that make the choices highly nuanced. Algorithm selection seems to be the most important for performance, though HPs should not be neglected (see Table \ref{tab:ms_h} for guidance). Default HPs (from packages or this paper) should be a reasonable starting point. For further tuning, one can consider prediction stability across HPs \citep{liuStabilityApproachRegularization2010,sunConsistentSelectionTuning2013,stroblAutomatedHyperparameterSelection2021}, though larger HP search spaces should be considered with care (risk of poor performances). Focusing on ``tunable'' HPs \citep{probst2019tunability} may help reduce the search space.\\
\noindent\textbf{Limitations.} This study is naturally limited by our choice of explored algorithms, HPs and simulation properties. It is worth noting, however, that we do not intend to identify the best possible learning algorithm or HPs. On the contrary, the objective of this work is to show that the appropriate algorithm choices are nuanced, as also recently shown in the treatment effect estimation domain \citep{curthSearchInsightsNot2023}, and that HPs should be part of that subtle decision-making process, further confirming the importance of HPs reported in the literature \citep{paineHyperparameterSelectionOffline2020,zhangImportanceHyperparameterOptimization2021,machlanskiHyperparameterTuningModel2023,tonshoffWhereDidGap2023}. And while more extensive search spaces are unlikely to negate such conclusions, it is worth pointing out that in our experiments we explore discrete HP values of continuous search spaces. As for the choice of HPs and values, we believe our setup accurately reflects common practice.\\
\noindent\textbf{Future work.} As increasing the sample size increases the proportion of good performances across HPs, a next step would be to examine if HP misspecification vanishes with infinite data. More challenging ``anterial graphs'' \citep{sadeghi2016marginalization} with cycles and undirected edges could be another direction. The surprising effectiveness of fixed HPs might be also worth a systematic study, with an emphasis on transfer between DGPs. Further study into the source of robustness to misspecified HPs is also in order (HP cardinality or algorithms). Furthermore, making advanced tuning metrics more accessible could help facilitate better practice. Finally, although some HP tuning metrics are general enough to perform algorithm selection (e.g. \citet{liuStabilityApproachRegularization2010,bizaOutofSampleTuningCausal2022}), doing so on real-life datasets is still a challenge. One promising direction could be validation that incorporates domain knowledge \citep{grunbaumQuantitativeProbingValidating2023}.

\acks{We would like to thank the reviewers for their insightful comments and suggestions that helped us improve this paper. This research was supported by the ESRC Research Centre on Micro-Social Change (MiSoC) -- ES/S012486/1, and in part through computational resources provided by the Business and Local Government Data Research Centre BLG DRC (ES/S007156/1) funded by the Economic and Social Research Council (ESRC).}

\bibliography{mylib}

\appendix

\newpage
\section{Experimental Details}

\subsection{Hyperparameters}\label{app:hyper}
We attempted to explore the hyperparameters as thoroughly as possible to make our findings general enough while at the same time managing computational demands that increase with each added hyperparameter and explored value. In addition, some methods are considerably more demanding than others, which makes the exploration even more difficult. With these constraints in mind, we believe our hyperparameter exploration accurately reflects common practice.

\begin{table}[!htb]
    \caption{Hyperparameter search spaces defined per algorithm. Note that hyperparameters are in most cases continuous, but we explore a discrete space of values.}
    \label{tab:hyper}
    \centering
    \resizebox{\textwidth}{!}{
    \begin{tabular}{lll}
\toprule
   algorithm & hyperparameters and values & card$^{\mathrm{a}}$ (total) \\
\midrule
    ANM & $alpha \in \{ 0.001^*, 0.002, 0.005, 0.01, 0.02, 0.05, 0.1, 0.2, 0.3, 0.5 \}$ & $10$ $(10)$ \\
    \midrule
    CAM & $cutoff \in \{ 0.001^*, 0.002, 0.005, 0.01, 0.02, 0.05, 0.1, 0.2, 0.3, 0.5 \}$ & $10$ $(10)$ \\
    {} & $score = $ nonlinear \\
    {} & $selmethod = $ gamboost \\
    {} & $prunmethod = $ gam \\
    \midrule
    FCI & $alpha \in \{ 0.001^*, 0.002, 0.005, 0.01, 0.02, 0.05, 0.1, 0.2, 0.3, 0.5 \}$ & $10$ $(10)$ \\
    {} & $test = $ fisher-z-test \\
    \midrule
    FGES & $penaltyDiscount \in \{ 0.001, 0.01, 0.1, 0.25, 0.5, 0.75, 1, 1.5^* \}$ & $8$ $(8)$ \\
    {} & $score = $ sem-bic \\
    \midrule
    LiNGAM & $thresh \in \{ 0.001, 0.002, 0.005, 0.01, 0.02, 0.05, 0.1, 0.2, 0.3, 0.5^* \}$ & $10$ $(20)$ \\
    {} & $max\_iter \in \{ 100^*, 1000 \}$ & $2$ \\
    \midrule
    NOTEARS & $lambda1 \in \{ 0.001, 0.002, 0.005, 0.01, 0.02, 0.05, 0.1, 0.2^*, 0.3, 0.5 \}$ & $10$ $(200)$ \\
    {} & $max\_iter \in \{ 100^*, 1000 \}$ & $2$ \\
    {} & $w\_threshold \in \{ 0.001, 0.002, 0.005, 0.01, 0.02, 0.05, 0.1, 0.2^*, 0.3, 0.5 \}$ & $10$ \\
    {} & $loss\_type = $ l2 \\
    {} & $h\_tol = 1e-8$ \\
    {} & $rho\_max = 1e+16$ \\
    \midrule
    NOTEARS MLP & $lambda1 \in \{ 0.001, 0.01^*, 0.1 \}$ & $3$ $(81)$ \\
    {} & $lambda2 \in \{ 0.001, 0.01, 0.1^* \}$ & $3$ \\
    {} & $w\_threshold \in \{ 0.1, 0.3, 0.5^* \}$ & $3$ \\
    {} & $hidden\_layers = 1$ \\
    {} & $hidden\_units \in \{ 8, 16^*, 32 \}$ & $3$ \\
    {} & $max\_iter = 100$ \\
    {} & $h\_tol = 1e-8$ \\
    {} & $rho\_max = 1e+16$ \\
    \midrule
    PC & $alpha \in \{0.001, 0.002^*, 0.005, 0.01, 0.02, 0.05, 0.1, 0.2, 0.3, 0.5\}$ & $10$ $(10)$ \\
    {} & $indepTest = $ gaussCItest \\
\bottomrule
\multicolumn{2}{l}{$^*$Found to perform best on average across all simulations (\textit{sim\_mean}).}\\
\multicolumn{2}{l}{$^{\mathrm{a}}$Cardinality of the hyperparameters considered in the experiments.}
\end{tabular}
}
\end{table}

\newpage
\subsection{Summary of Algorithms}\label{app:algs}
This study incorporates the following algorithms and hyperparameters.

\begin{itemize}
    \item \textbf{PC} \citep{spirtesAlgorithmFastRecovery1991}. Peter and Clark algorithm. Constraint-based approach that starts with a fully-connected undirected graph and removes edges based on conditional independence tests. Next, it attempts to orient as many of the remaining edges as possible. The result is a CPDAG.\\ \textbf{Hyperparameters:} \textit{alpha} (significance level for conditional independence tests).
    \item \textbf{FCI} \citep{Spirtes1993}. Fast Causal Inference. Constraint-based. An important generalisation of PC to unknown confounding variables.\\ \textbf{Hyperparameters:} \textit{alpha} (significance level for conditional independence tests).
    \item \textbf{FGES} \citep{ramseyMillionVariablesMore2017}. Fast Greedy Equivalence Search. Optimised and parallelised version of the original score-based GES algorithm \citep{chickeringOptimalStructureIdentification2002}. It starts with an empty graph and adds an edge that yields maximum score improvement until no significant score gain is achieved. Then it removes edges in the same greedy manner until a plateau.\\ \textbf{Hyperparameters:} \textit{penaltyDiscount} (sparsity penalty).
    \item \textbf{LiNGAM} \citep{shimizuLinearNonGaussianAcyclic2006}. Linear Non-Gaussian Acyclic Model. Assumes linear SEMs and non-Gaussian noise that enters additively: $X_j = \sum_{k \in \paj} w_{jk} X_k + \epsilon_j$.\\ \textbf{Hyperparameters:} \textit{max\_iter} (FastICA \citep{fastICA}), \textit{thresh} (post-prunning threshold).
    \item \textbf{ANM} \citep{hoyerNonlinearCausalDiscovery2008}. Additive Noise Model. Assumes nonlinear SEMs and additive noise: $X_j = f_j(X_{\paj}) + \epsilon_j$.\\ \textbf{Hyperparameters:} \textit{alpha} (significance level for the independence test).
    \item \textbf{CAM} \citep{buhlmannCAMCausalAdditive2014}. Causal Additive Models. Assumes a generalised additive noise model with additive noise and functions: $X_j = \sum_{k \in \paj} f(X_k) + \epsilon_j$.\\ \textbf{Hyperparameters:} \textit{cutoff} (variable selection threshold).
    \item \textbf{NOTEARS} \citep{zhengDAGsNOTEARS2018}. Score-based continuous DAG optimisation with a smooth acyclicity regularisation term. Assumes linear SEMs with additive noise.\\ \textbf{Hyperparameters:} \textit{lambda1} (sparsity term), \textit{max\_iter} (optimisation steps) and \textit{w\_threshold} (post-prunning threshold).
    \item \textbf{NOTEARS MLP} \citep{zhengLearningSparseNonparametric2020}. Nonlinear extension of \textit{NOTEARS} by incorporating the Multi-Layer Perceptron (MLP). Assumes nonlinear SEMs with additive noise.\\ \textbf{Hyperparameters:} \textit{lambda1} (sparsity term), \textit{lambda2} (regularisation strength), \textit{w\_threshold} (post-prunning threshold), \textit{hidden\_units} (number of units in the hidden layer).
\end{itemize}

Many traditional algorithms, such as PC, FCI and FGES, make the standard set of assumptions that involve sufficiency, faithfulness and Markov condition. These, however, are often not enough to identify a unique DAG as a solution, which is a major drawback of these methods (they output CPDAGs). Making assumptions about distributions and functional forms of the data generating process seems to be critical to overcome this issue (all methods above except for PC, FCI and FGES output DAGs).

\begin{table}[!htb]
    \caption{Summary of incorporated algorithms and their sources. Recommended default hyperparameters have been derived from respective papers as much as possible. If necessary, they have been further supplemented with defaults suggested within respective packages.}
    \label{tab:summary_algs}
    \centering
    \begin{tabular}{llcc}
\toprule
   algorithm & default hyperparameters & package & paper \\
\midrule
    ANM & $alpha = 0.05$ & gCastle & \citep{hoyerNonlinearCausalDiscovery2008} \\
    \midrule
    CAM & $cutoff = 0.001$ & cdt & \citep{buhlmannCAMCausalAdditive2014} \\
    {} & $score = $ nonlinear & {} & {} \\
    {} & $selmethod = $ gamboost & {} & {} \\
    {} & $prunmethod = $ gam & {} & {} \\
    \midrule
    FCI & $alpha = 0.01$ & tetrad & \citep{Spirtes1993} \\
    \midrule
    FGES & $penaltyDiscount = 2.0$ & tetrad & \citep{ramseyMillionVariablesMore2017} \\
    \midrule
    LiNGAM & $thresh = 0.3$ & gCastle & \citep{shimizuLinearNonGaussianAcyclic2006} \\
    {} & $max\_iter = 1000$ & {} & {} \\
    \midrule
    NOTEARS & $lambda1 = 0.1$ & gCastle & \citep{zhengDAGsNOTEARS2018} \\
    {} & $max\_iter = 100$ & {} & {} \\
    {} & $w\_threshold = 0.3$ & {} & {} \\
    {} & $loss\_type = $ l2 & {} & {} \\
    {} & $h\_tol = 1e-8$ & {} & {} \\
    {} & $rho\_max = 1e+16$ & {} & {} \\
    \midrule
    NOTEARS MLP & $lambda1 = 0.01$ & gCastle & \citep{zhengLearningSparseNonparametric2020} \\
    {} & $lambda2 = 0.01$ & {} & {} \\
    {} & $w\_threshold = 0.3$ & {} & {} \\
    {} & $hidden\_layers = 1$ & {} & {} \\
    {} & $hidden\_units = 10$ & {} & {} \\
    {} & $max\_iter = 100$ & {} & {} \\
    {} & $h\_tol = 1e-8$ & {} & {} \\
    {} & $rho\_max = 1e+16$ & {} & {} \\
    \midrule
    PC & $alpha = 0.01$ & pcalg & \citep{spirtesAlgorithmFastRecovery1991} \\
\bottomrule
\end{tabular}
\end{table}

\newpage
\subsection{Summary of Packages}

\begin{table}[!htbp]
    \caption{Summary of incorporated algorithm packages.}
    \label{tab:packages}
    \centering
    \begin{tabular}{lcl}
    \toprule
    package & paper & link \\
    \midrule
         gCastle & \citep{zhang2021gcastle} & { \tiny \url{https://github.com/huawei-noah/trustworthyAI/tree/master/gcastle} }\\
         cdt & \citep{kalainathanCausalDiscoveryToolbox2020} & { \tiny \url{https://fentechsolutions.github.io/CausalDiscoveryToolbox} }\\
         tetrad & \citep{ramsey2018tetrad} & { \tiny \url{https://cmu-phil.github.io/tetrad/manual/}} \\
         pcalg & \citep{pcalg2012} & { \tiny \url{https://cran.r-project.org/package=pcalg}} \\
    \bottomrule
\end{tabular}
\end{table}

\subsection{Performance Evaluation}\label{app:eval}
We incorporate commonly used evaluation metrics that are provided via Benchpress \citep[appendix A.1.]{riosBenchpressScalableVersatile2022}. For the convenience of the reader, we briefly describe here those that are useful for this study.

\subsubsection{SHD}
Let us define $E$ and $E'$ as a set of edges of the true and predicted DAG respectively. Then, for $e \in E'$, true positives (TP) and false positives (FP) are assigned as follows:

\begin{equation}
TP(e)=\begin{cases}
1 & \text{ if } e \in E \text{ and correctly oriented} \\ 
0.5 & \text{ if } e \in E \text{ and incorrectly oriented} \\ 
0 & \text{ otherwise } 
\end{cases}    
\end{equation}

\begin{equation}
FP(e)=\begin{cases}
1 & \text{ if } e \notin E \\ 
0.5 & \text{ if } e \in E \text{ and incorrectly oriented} \\ 
0 & \text{ otherwise } 
\end{cases}    
\end{equation}
where TP and FP are sums of all TP(e) and FP(e) scores respectively. The \textit{structural Hamming distance} (SHD) aggregates the number additions, removals and reversals in predicted edges so they match the true ones ($E = E'$). It can be defined as:

\begin{equation}
    SHD = |E| - TP + FP
\end{equation}

Note the SHD defined as above allows to evaluate mixed graphs, that is, compare DAGs to CPDAGs. If, for instance, a predicted undirected edge exists in $E$ but is supposed to be directed, it will result in $TP=0.5$ and $FP=0.5$, ultimately leading to $SHD=1$. This shows that the need to orient an undirected edge is treated equally as the need to add, remove or reverse an edge so $E = E'$. Such evaluation puts algorithms outputting CPDAGs at a disadvantage compared to DAG-only methods. We justify it on the grounds that the main focus of this study is DAG recovery, hence any predicted undirected edge is treated as any other mistake. The ability to evaluate mixed graphs is an important feature for this study as it allows us to compare algorithms outputting CPDAGs and DAGs.

\subsubsection{False Positives and Negatives}
The \textit{false positives} (FPs) and \textit{false negatives} (FNs) that we use as standalone performance metrics differ from those defined as part of SHD above. Crucially, the FP and FN metrics we use are always computed based on graph skeletons. To achieve this, all directed edges of a predicted or true graph are converted to undirected ones. This modification makes the comparison between algorithms that output CPDAGs and DAGs more fair. Once the graphs in question are converted to skeletons, we can define the metrics as follows.

Let us define $E$ and $E'$ as a set of undirected edges of the true and predicted graph skeleton respectively. Then, for $e \in E$ and $e' \in E'$, false positives (FP) and false negatives (FN) are assigned as follows:

\begin{equation}
FP(e')=\begin{cases}
1 & \text{ if } e' \notin E \\ 
0 & \text{ otherwise } 
\end{cases}    
\end{equation}

\begin{equation}
FN(e)=\begin{cases}
1 & \text{ if } e \notin E' \\
0 & \text{ otherwise } 
\end{cases}    
\end{equation}
where FP and FN are sums of all FP(e') and FN(e) scores respectively.

\subsection{Code and Data}\label{app:code}
All numerical experiments can be fully replicated using the code and data that are available online at: {\footnotesize \url{https://github.com/misoc-mml/hyperparams-causal-discovery}}.

\newpage
\section{Supplemental Results}\label{app:results}
The following results complement the ones presented in the main content of the paper. Although they do not change the overall conclusions of the paper, they offer additional analysis that may facilitate a deeper understanding of the problem and the final outcomes.

\subsection{Best Performances}
\begin{figure}[htbp]
    \centering
    \subfigure[linear SEM with Gumbel noise]{\includegraphics[width=\textwidth]{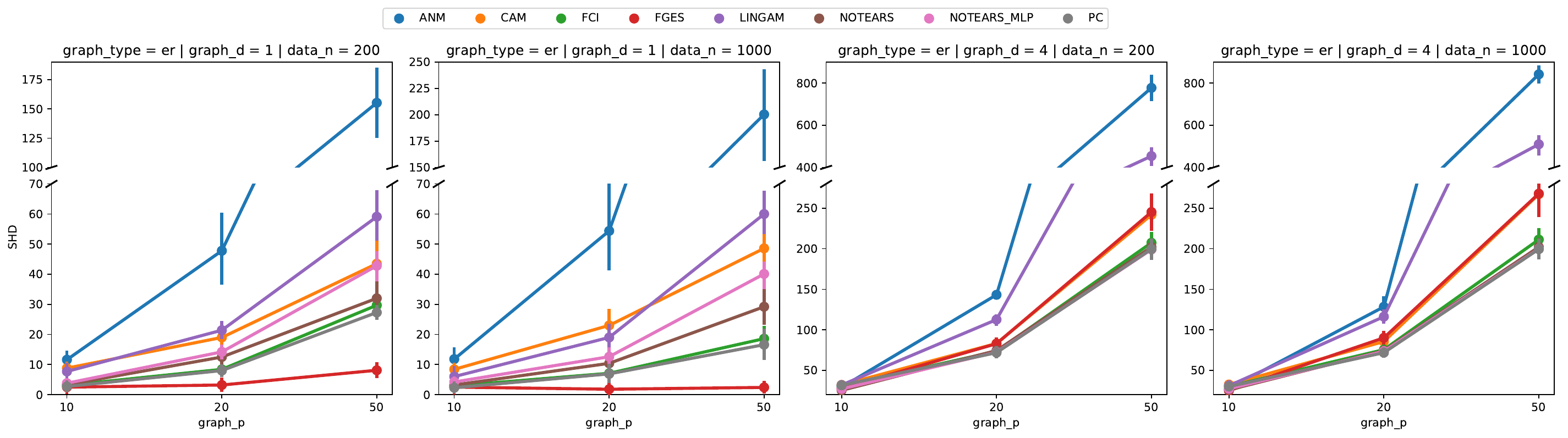}\label{fig:oracle_gumbel}}
    \subfigure[nonlinear (GP) SEM with Gaussian noise]{\includegraphics[width=\textwidth]{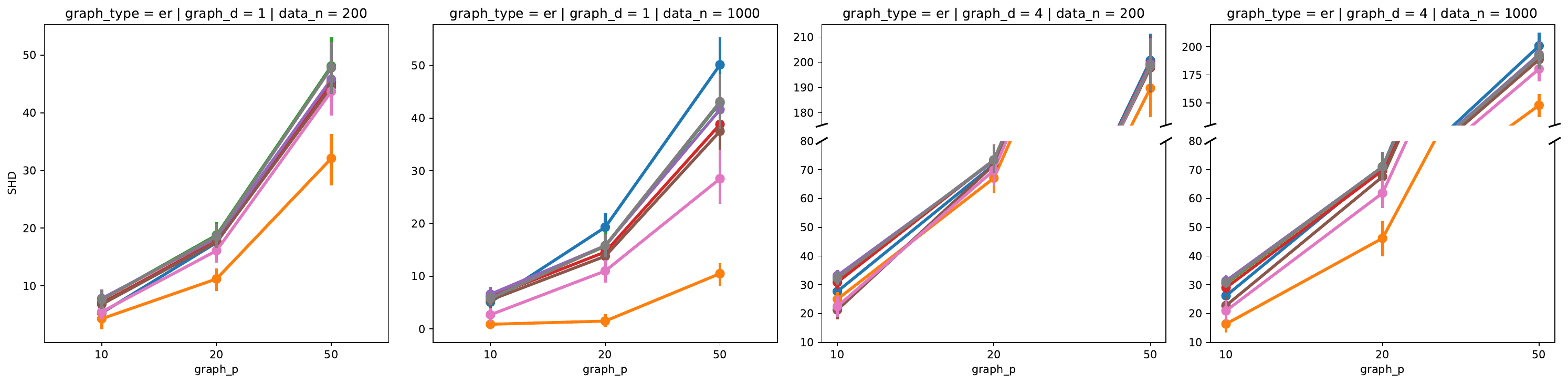}\label{fig:oracle_gp}}
    \caption{Best performances against ER graphs. Error bars are standard errors.}
    \label{fig:oracle}
\end{figure}

\begin{figure}[htbp]
    \centering
    \subfigure[linear SEM with Gumbel noise]{\includegraphics[width=\textwidth]{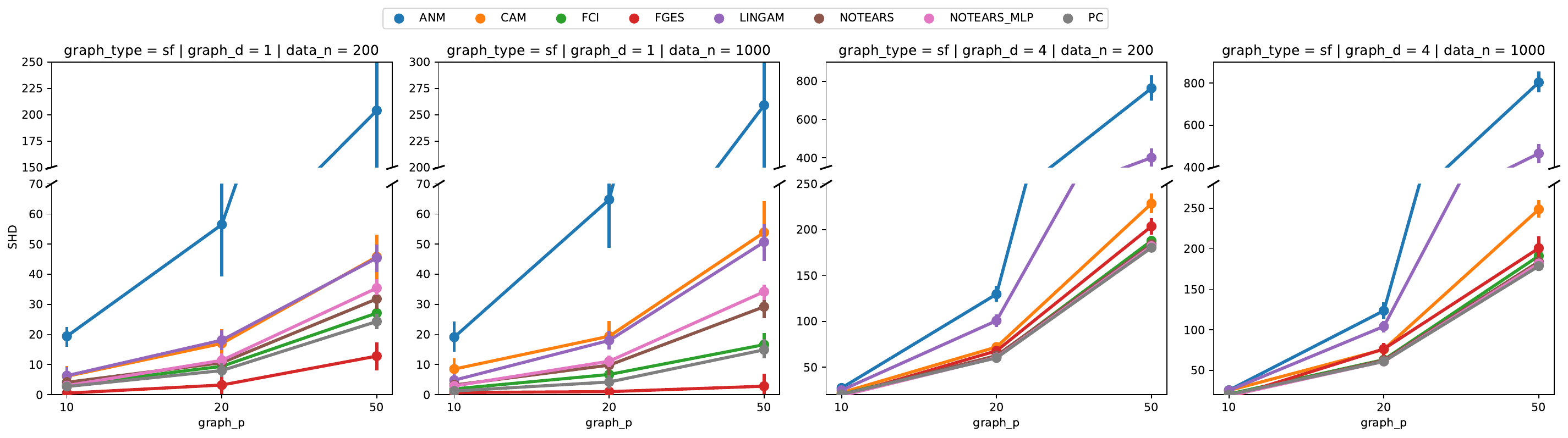}}
    \subfigure[nonlinear (GP) SEM with Gaussian noise]{\includegraphics[width=\textwidth]{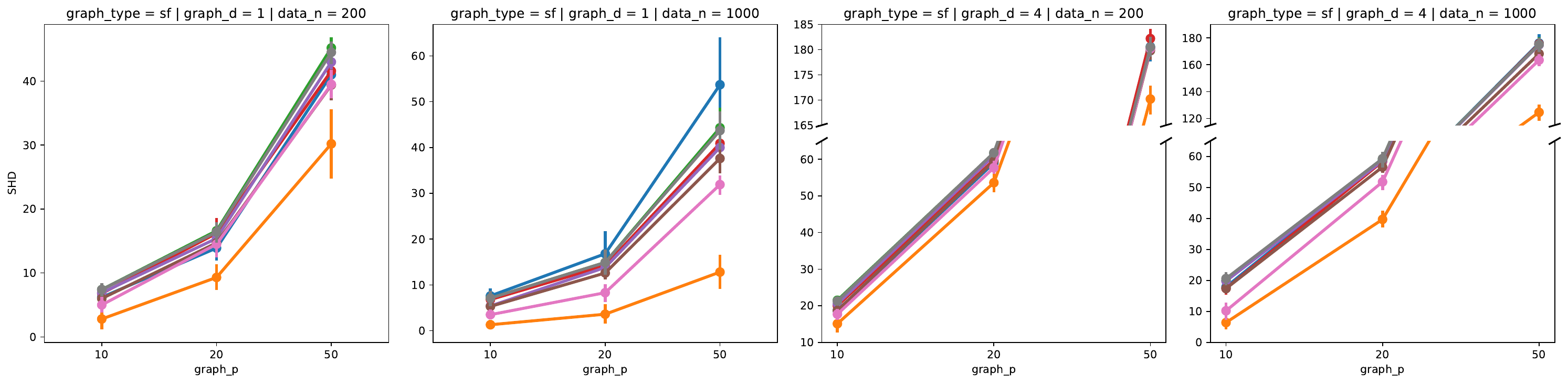}}
    \caption{Best performances against SF graphs. Error bars are standard errors.}
    \label{fig:oracle_sf}
\end{figure}

Figures \ref{fig:oracle} and \ref{fig:oracle_sf} focus on performances achieved specifically with the best (oracle) hyperparameters. These correspond to the best performances presented in Figure \ref{fig:hq} (blue bars). Some important observations: a) algorithms differ in performance even with access to a hyperparameter oracle, b) number of graph nodes and edge density can significantly impact performance, and c) no algorithm performs best under all conditions (linear vs. nonlinear).

\newpage
\subsection{Large Graphs with Large Sample Size}
\begin{figure}[htbp]
    \centering
    \subfigure[best performances]{\includegraphics[width=0.99\textwidth]{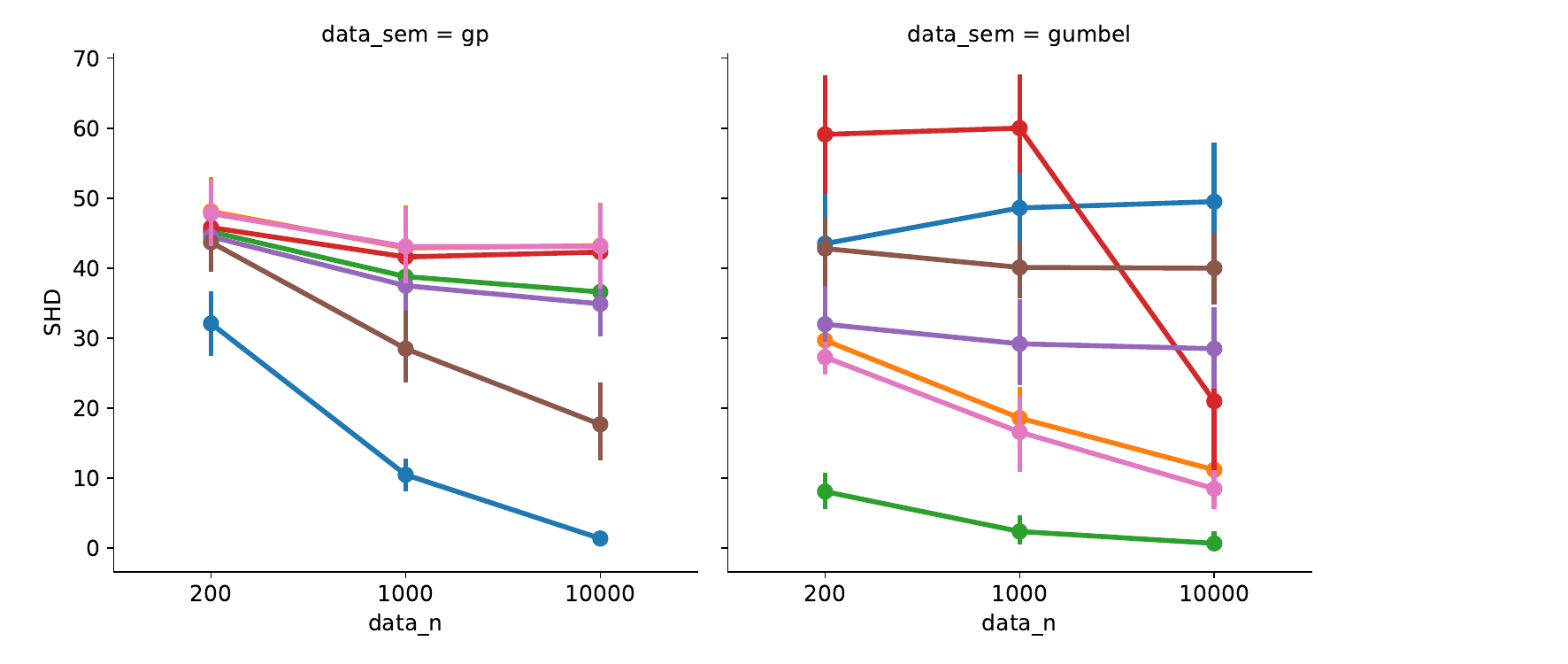}}
    \subfigure[worst performances]{\includegraphics[width=0.99\textwidth]{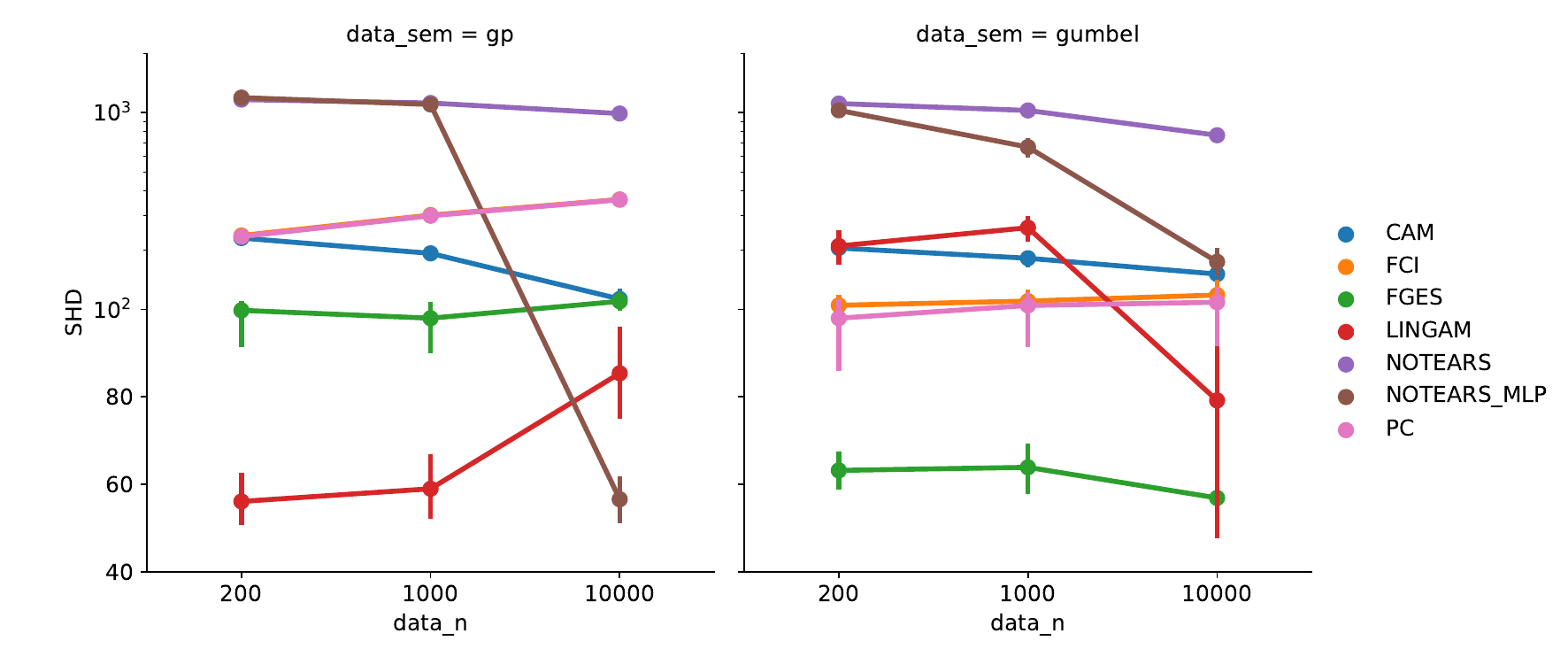}}
    \caption{Performances for ER1 $p=50$ graphs.}
    \label{fig:oracle_10k}
\end{figure}

Previous results showed that $p=50$ graphs are much more challenging than smaller ones. Figure \ref{fig:oracle_10k}a demonstrates that with enough data samples, even such larger graphs are possible to solve accurately. As presented in the subfigure (b), increased sample size can also help with robustness. Note also how the degree of the benefits vary between algorithms.

\begin{figure}[htbp]
    \centering
    \includegraphics[width=\textwidth]{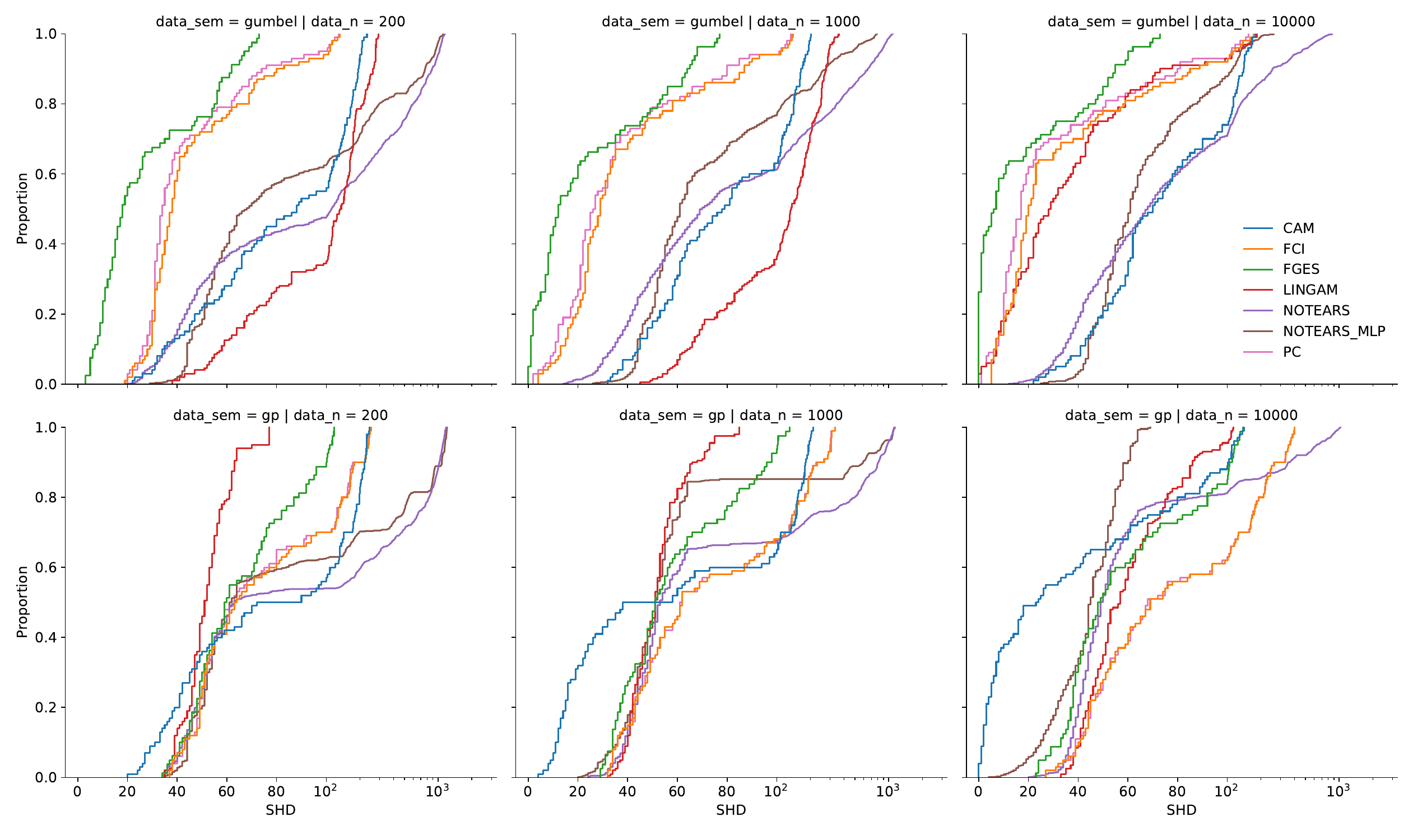}
    \caption{Distribution of SHD performances across all hyperparameters (ER1 $p=50$ graphs).}
    \label{fig:10k_shd}
\end{figure}

\begin{figure}[htbp]
    \centering
    \includegraphics[width=\textwidth]{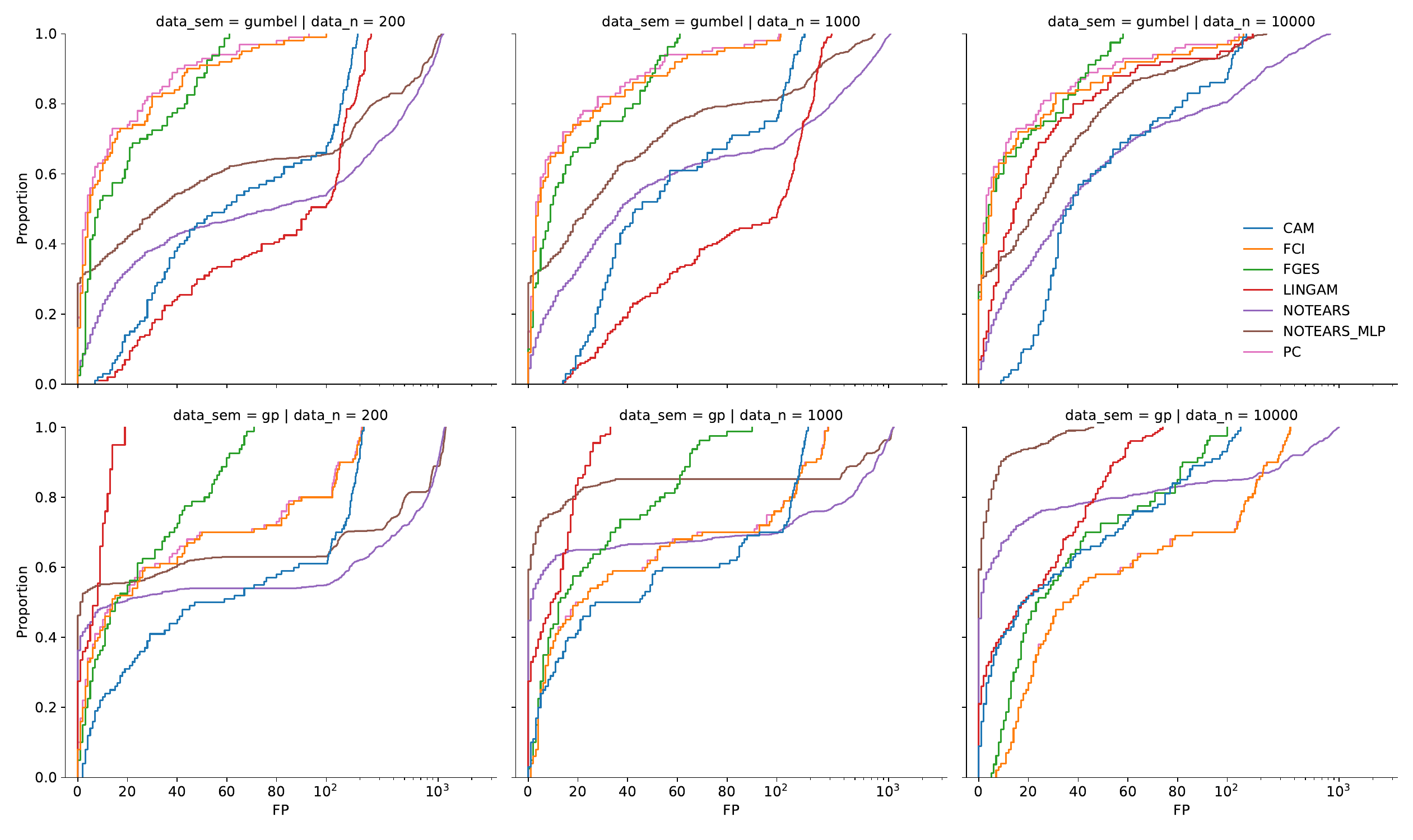}
    \caption{Distribution of false positives (FPs) across all hyperparameters (ER1 $p=50$ graphs).}
    \label{fig:10k_fp}
\end{figure}

\begin{figure}[htbp]
    \centering
    \includegraphics[width=\textwidth]{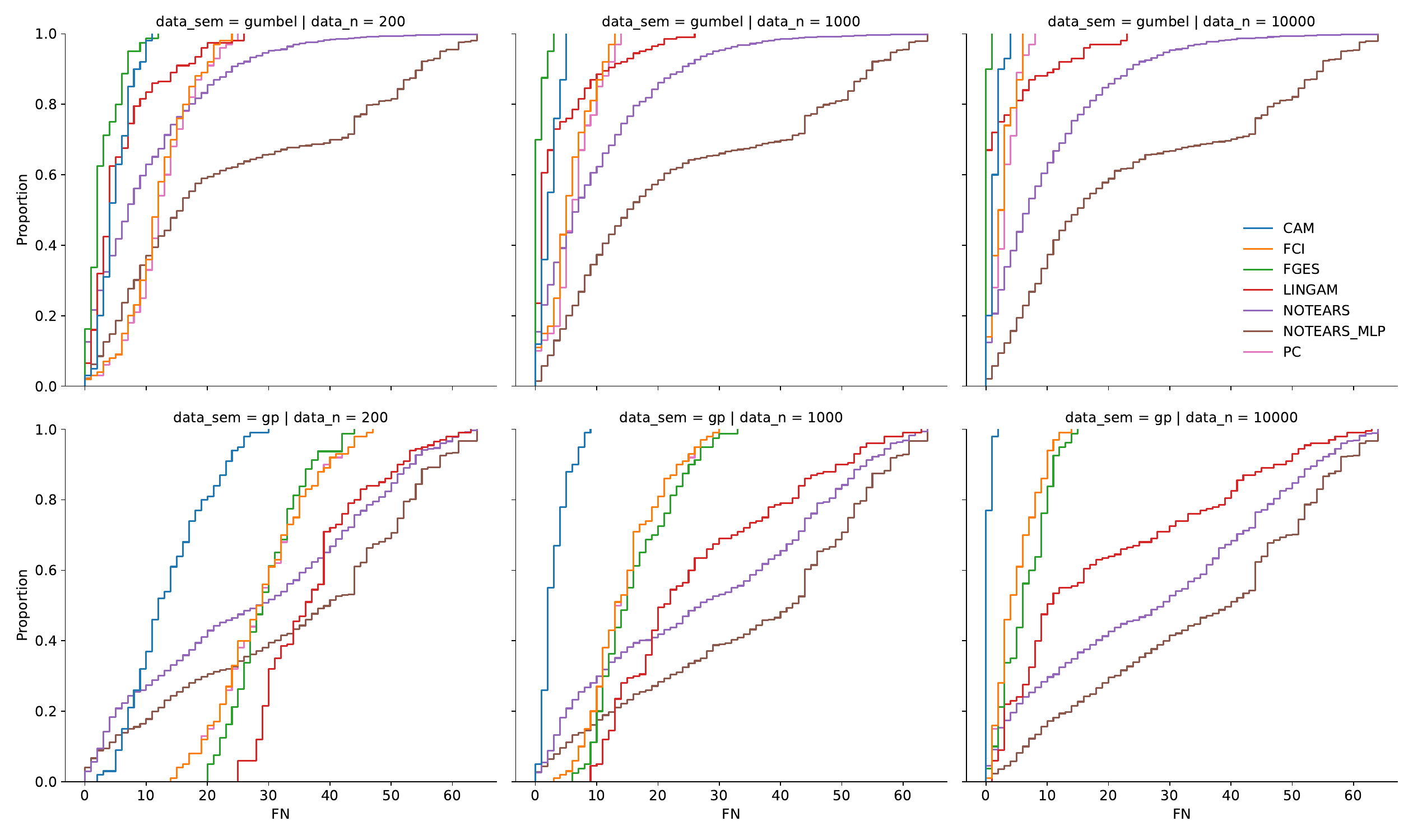}
    \caption{Distribution of false negatives (FNs) across all hyperparameters (ER1 $p=50$ graphs).}
    \label{fig:10k_fn}
\end{figure}

Figures \ref{fig:10k_shd}, \ref{fig:10k_fp} and \ref{fig:10k_fn} further extend the results to performance distributions over all hyperparameters (SHD, FP, and FN respectively). Notice how larger sample size increases the proportion of good performances (the lines shift to the left and hit higher proportion numbers for lower metric values). If the improvement trend remains for even larger sample sizes, one can wonder if the HP misspecification issue could be solved entirely by larger quantities of data alone.

\newpage
\subsection{Performance vs. Hyperparameter Quality}
\begin{figure}[htbp]
    \centering
    \includegraphics[width=\textwidth]{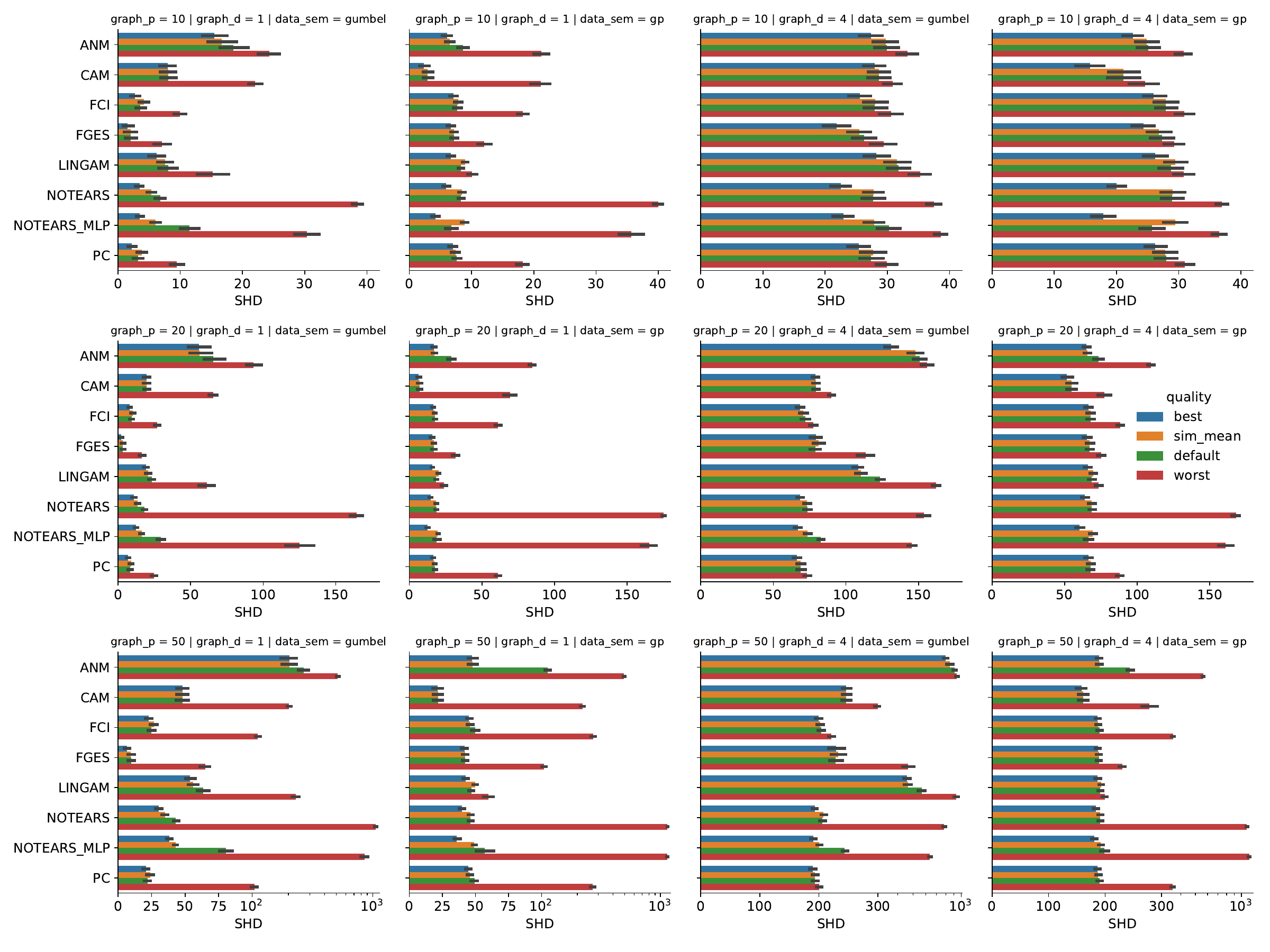}
    \caption{SHD performances depending on the quality of selected hyperparameters (colours), grouped by DGP properties such as number of nodes (\textit{graph\_p}), edge density (\textit{graph\_d}) and SEM type (\textit{data\_sem}; gumbel is linear, gp nonlinear).}
    \label{fig:hq_shd}
\end{figure}

\begin{figure}[htbp]
    \centering
    \includegraphics[width=\textwidth]{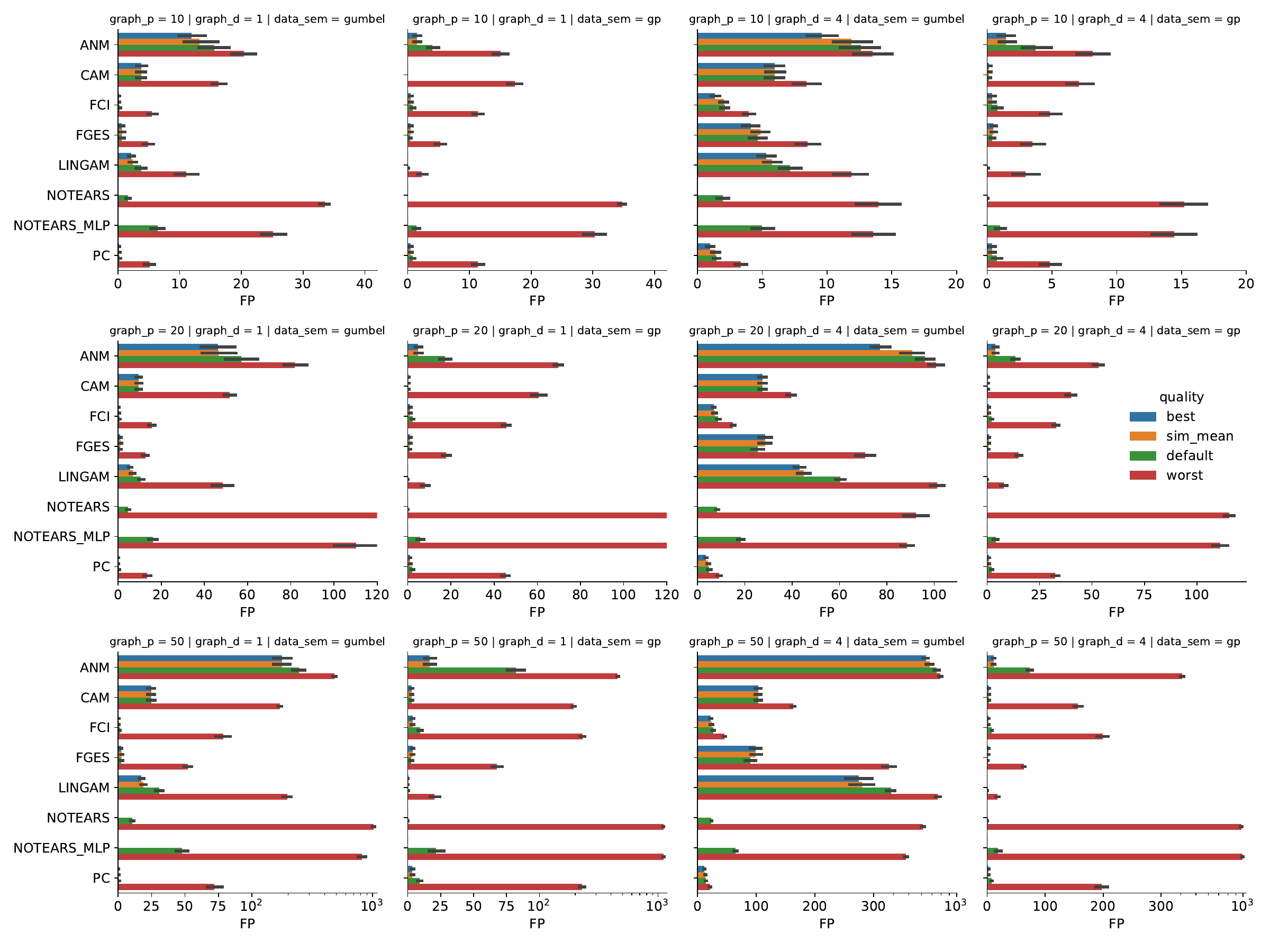}
    \caption{FP (false positive) performances depending on the quality of selected hyperparameters (colours), grouped by DGP properties such as number of nodes (\textit{graph\_p}), edge density (\textit{graph\_d}) and SEM type (\textit{data\_sem}; gumbel is linear, gp nonlinear).}
    \label{fig:hq_fp}
\end{figure}

\begin{figure}[htbp]
    \centering
    \includegraphics[width=\textwidth]{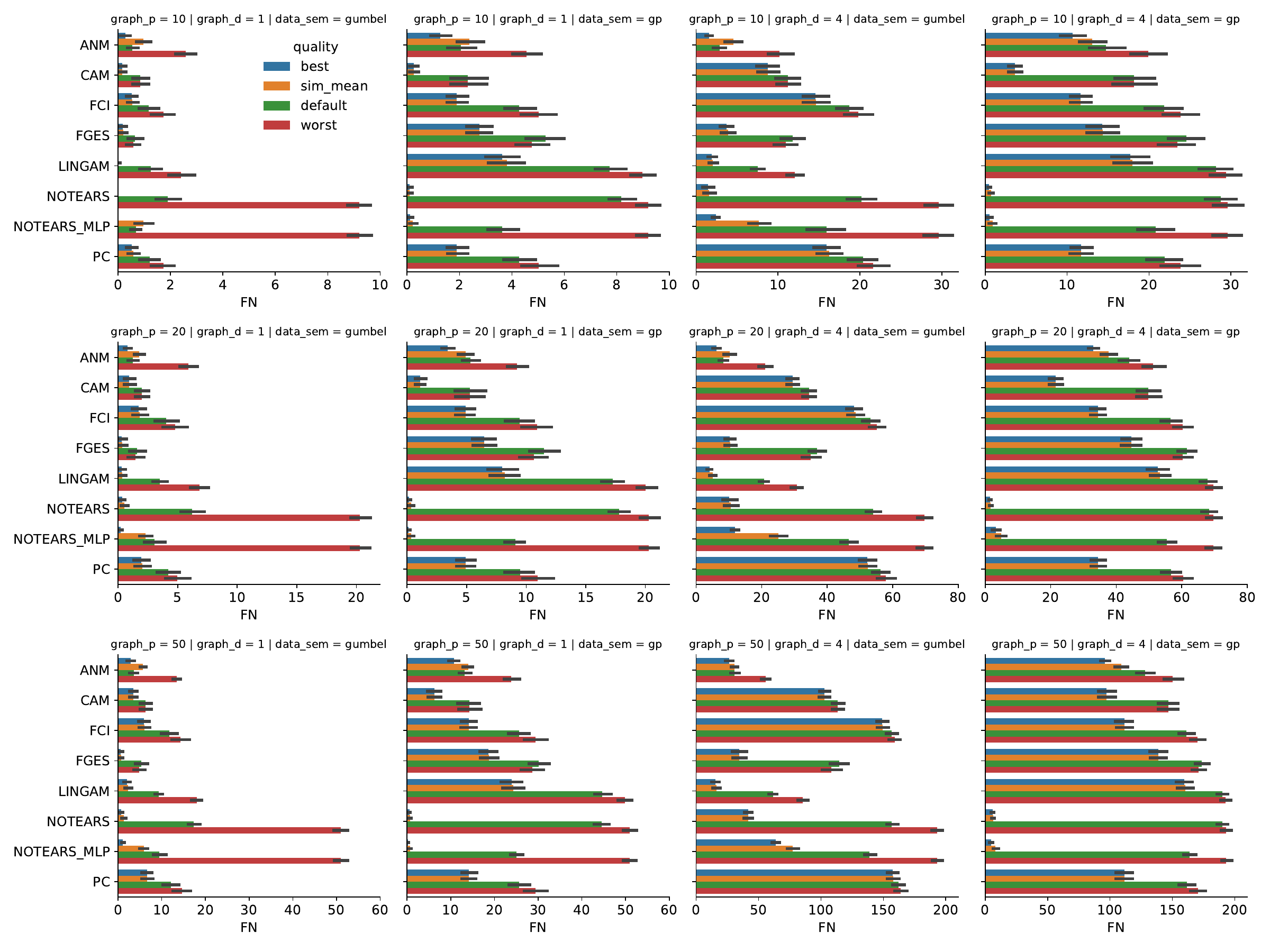}
    \caption{FN (false negative) performances depending on the quality of selected hyperparameters (colours), grouped by DGP properties such as number of nodes (\textit{graph\_p}), edge density (\textit{graph\_d}) and SEM type (\textit{data\_sem}; gumbel is linear, gp nonlinear).}
    \label{fig:hq_fn}
\end{figure}

Figures \ref{fig:hq_shd}, \ref{fig:hq_fp} and \ref{fig:hq_fn} complement Figure \ref{fig:hq} from the main content by showing the results for other types of DGPs.

\newpage
\subsection{Performance Distribution Across Hyperparameters}
\begin{figure}[htbp]
    \centering
    \subfigure[\textit{data\_sem}]{\includegraphics[width=0.49\textwidth]{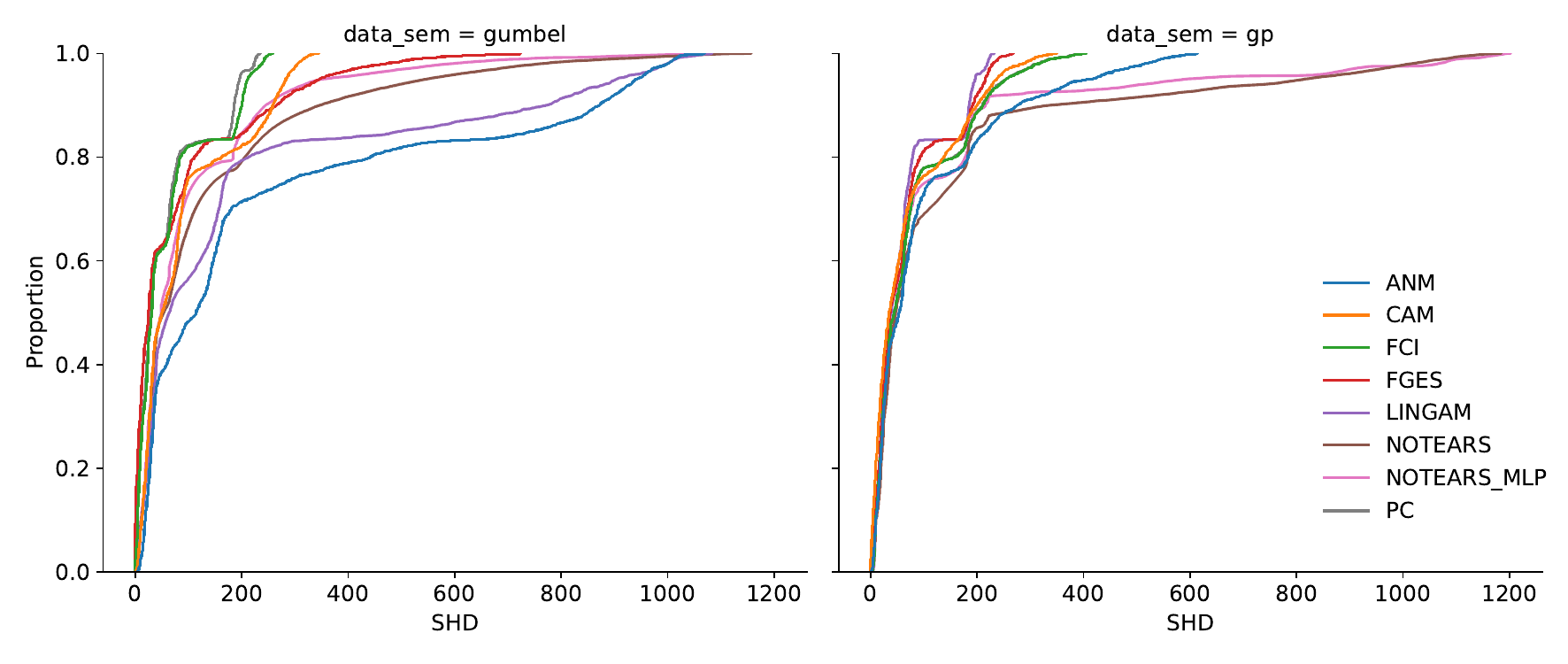}}
    \subfigure[\textit{data\_n}]{\includegraphics[width=0.49\textwidth]{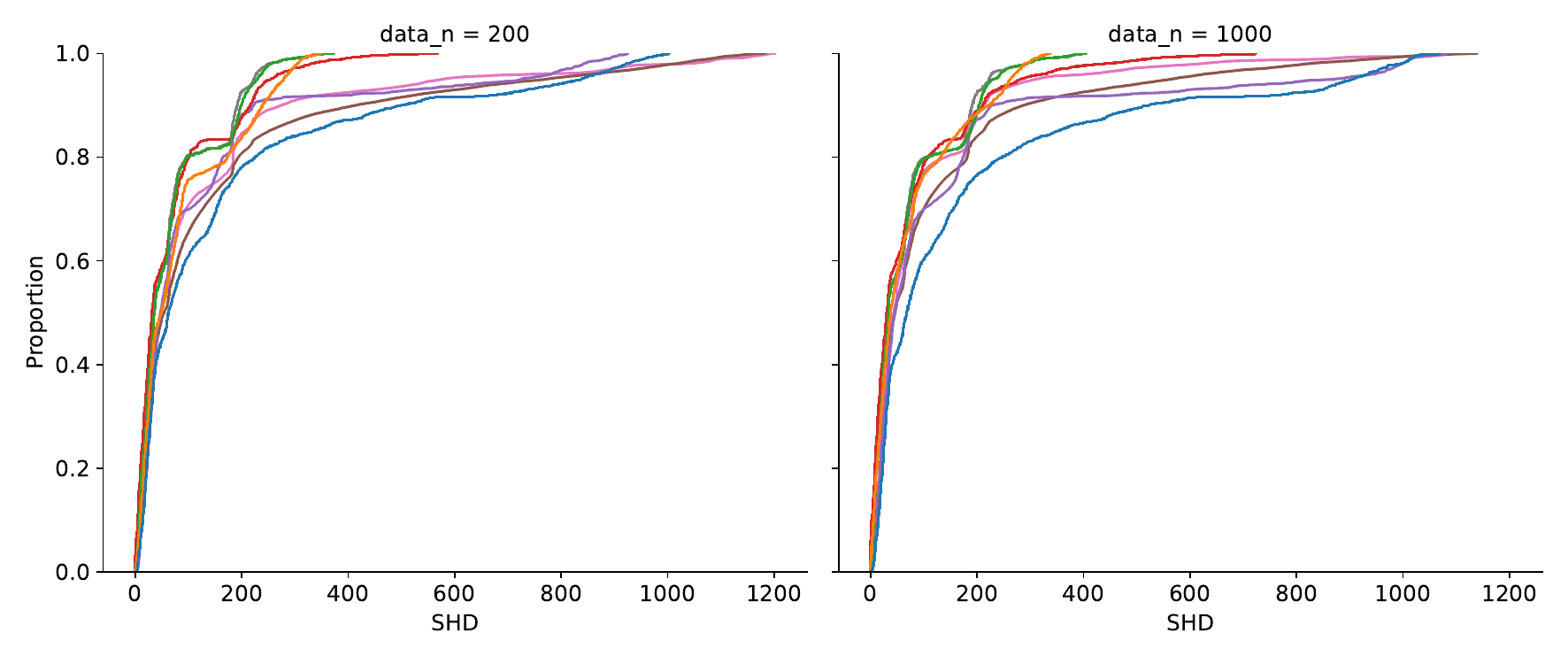}}
    \subfigure[\textit{graph\_d}]{\includegraphics[width=0.49\textwidth]{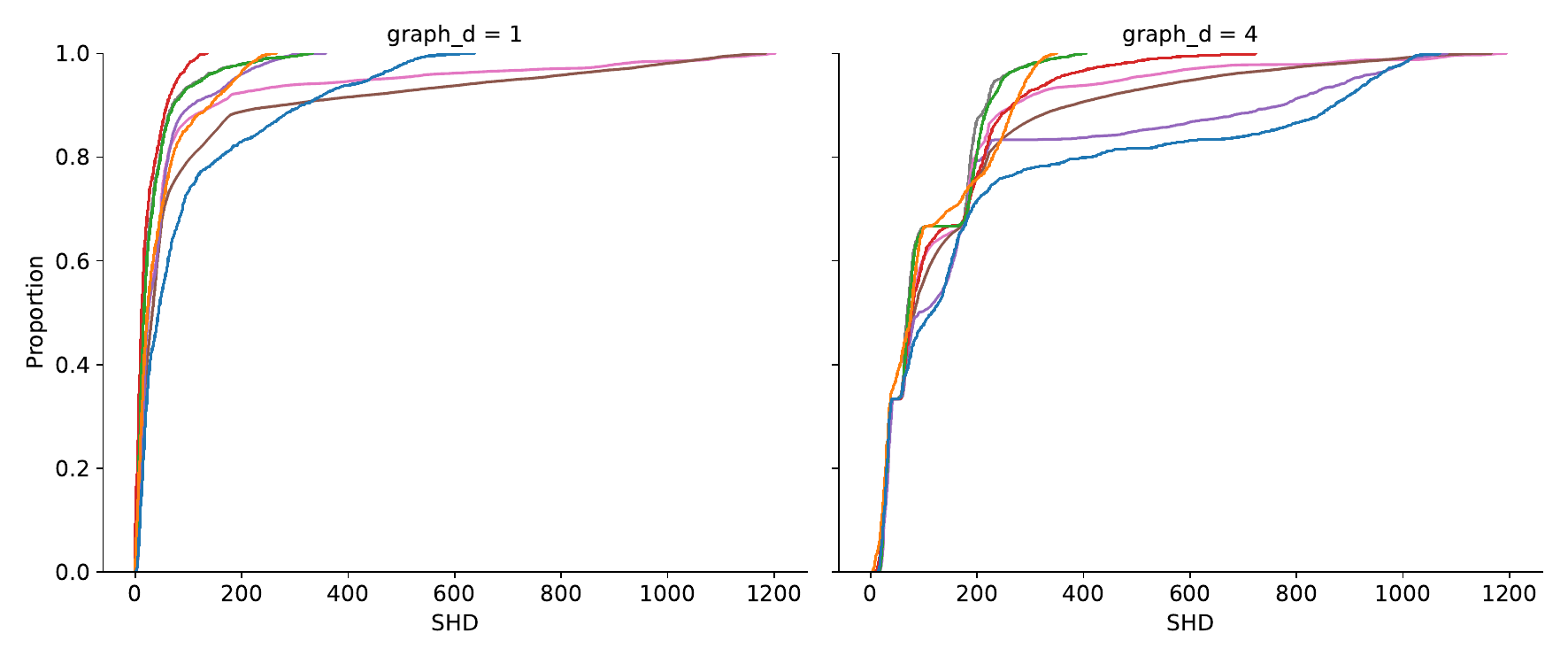}}
    \subfigure[\textit{graph\_type}]{\includegraphics[width=0.49\textwidth]{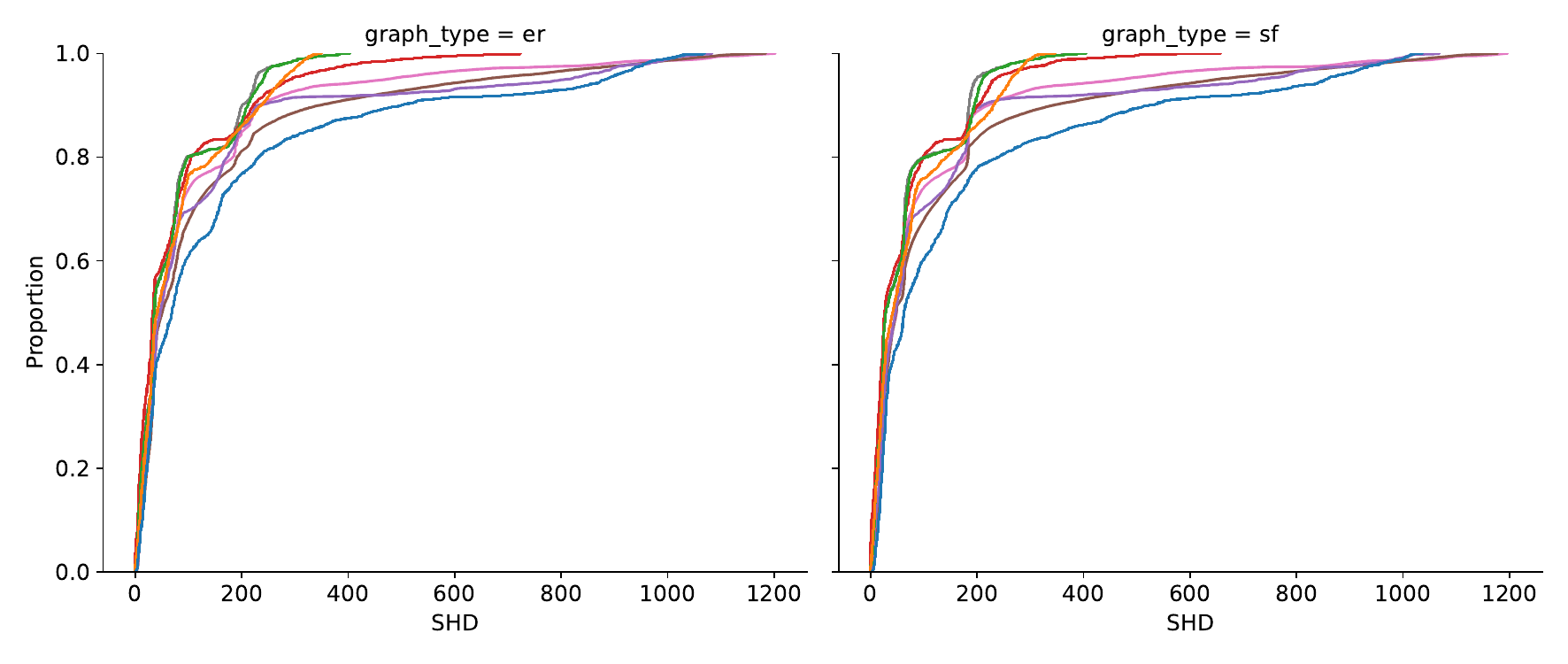}}
    \subfigure[\textit{graph\_p}]{\includegraphics[width=0.7\textwidth]{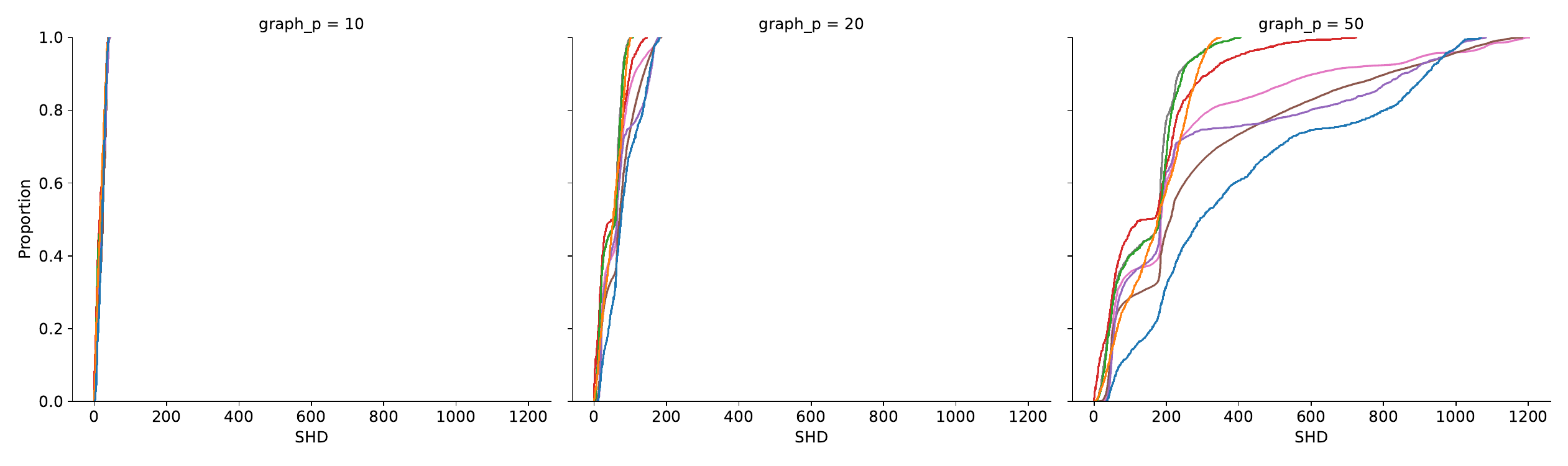}}
    \caption{Distributions of SHD performances across all hyperparameters, grouped by SEM types (\textit{data\_sem}), sample size (\textit{data\_n}), edge density (\textit{graph\_d}), graph type (ER or SF) and number of nodes (\textit{graph\_p}).}
    \label{fig:dist_shd}
\end{figure}

\begin{figure}[htbp]
    \centering
    \subfigure[\textit{data\_sem}]{\includegraphics[width=0.49\textwidth]{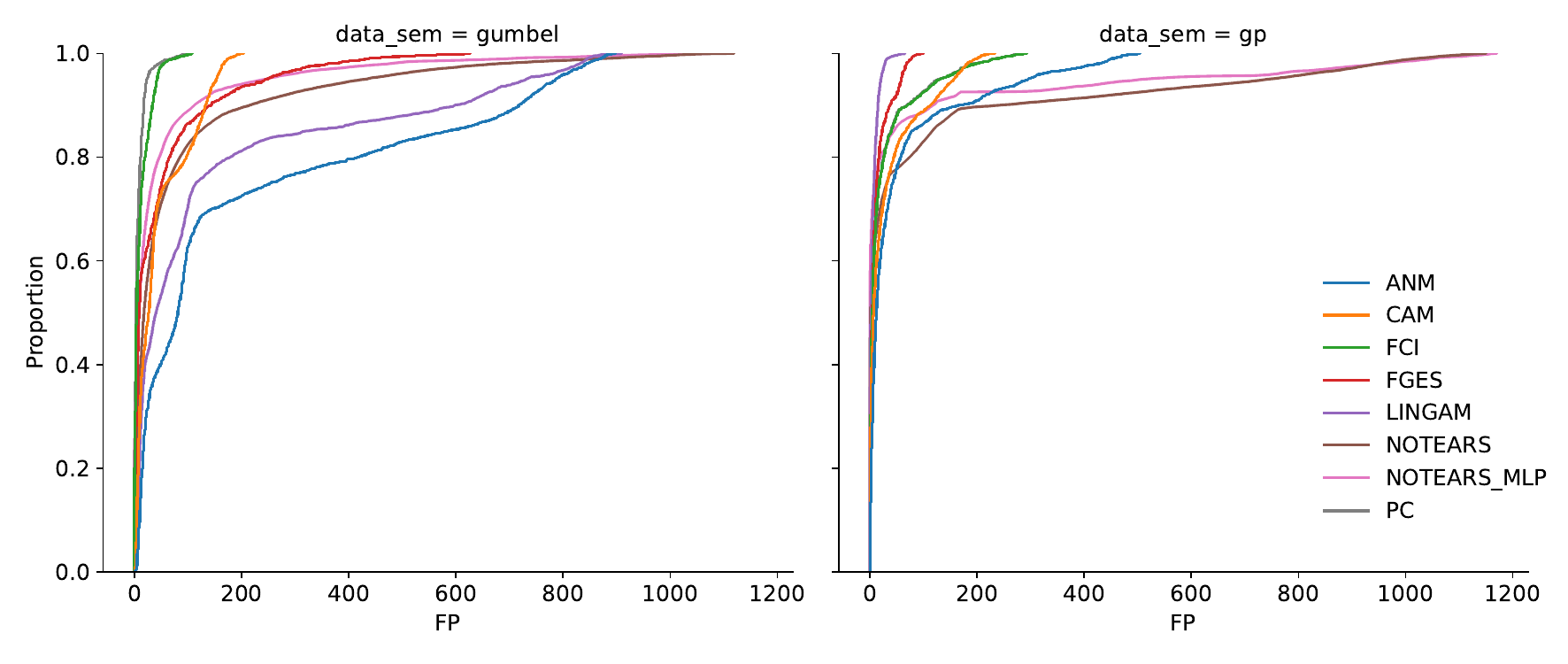}}
    \subfigure[\textit{data\_n}]{\includegraphics[width=0.49\textwidth]{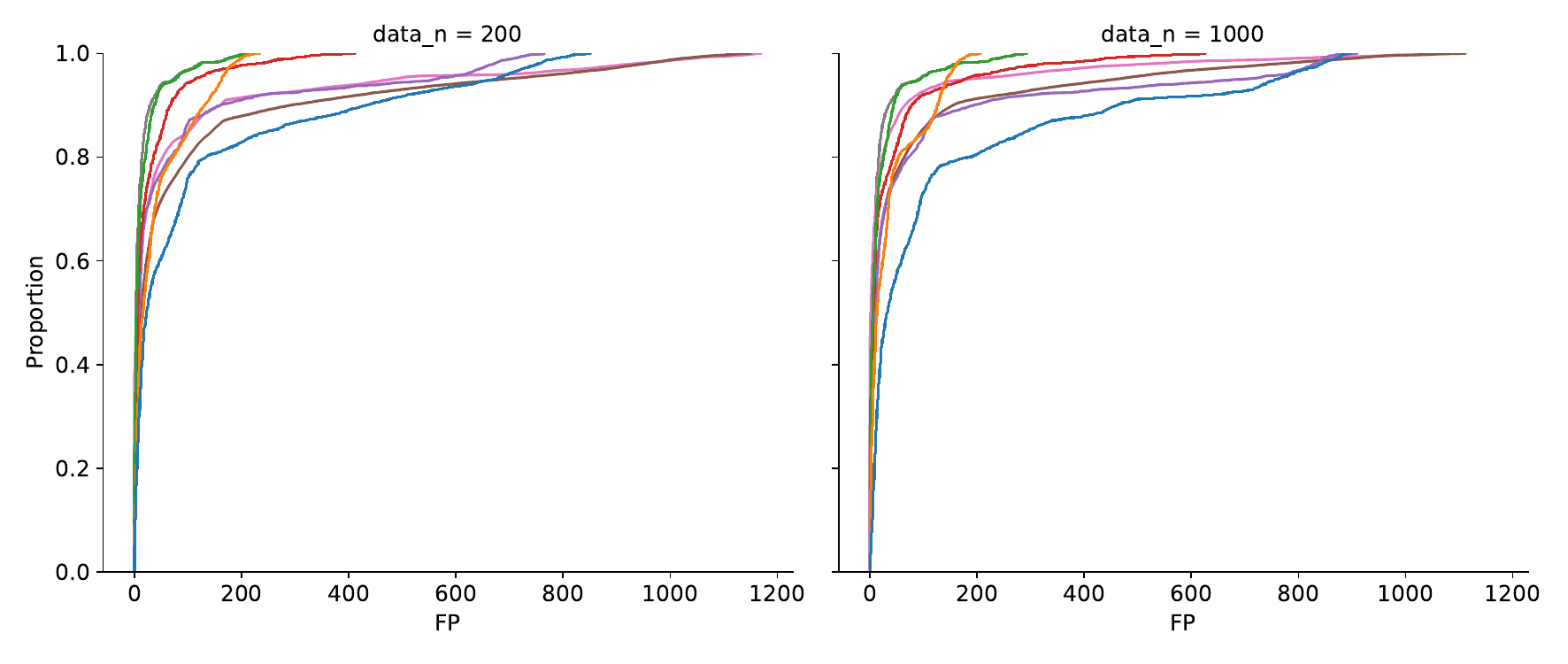}}
    \subfigure[\textit{graph\_d}]{\includegraphics[width=0.49\textwidth]{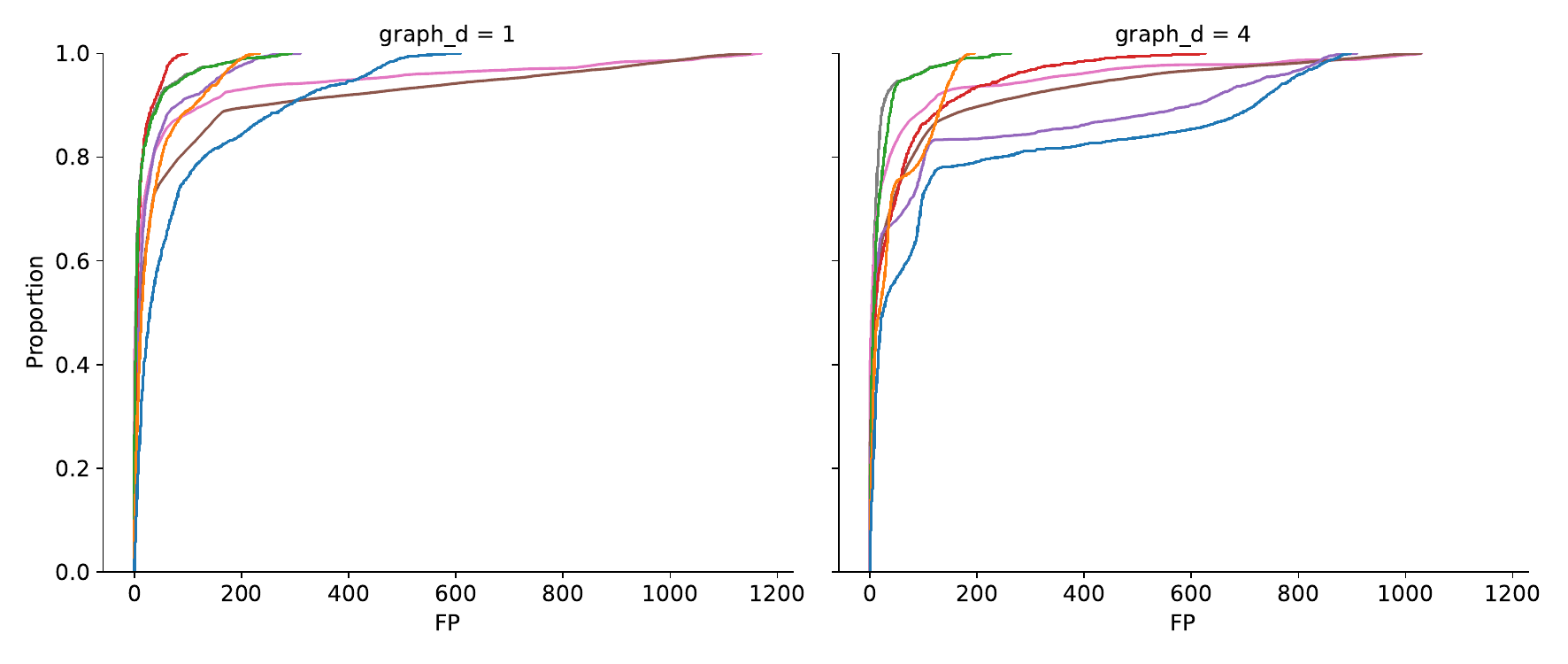}}
    \subfigure[\textit{graph\_type}]{\includegraphics[width=0.49\textwidth]{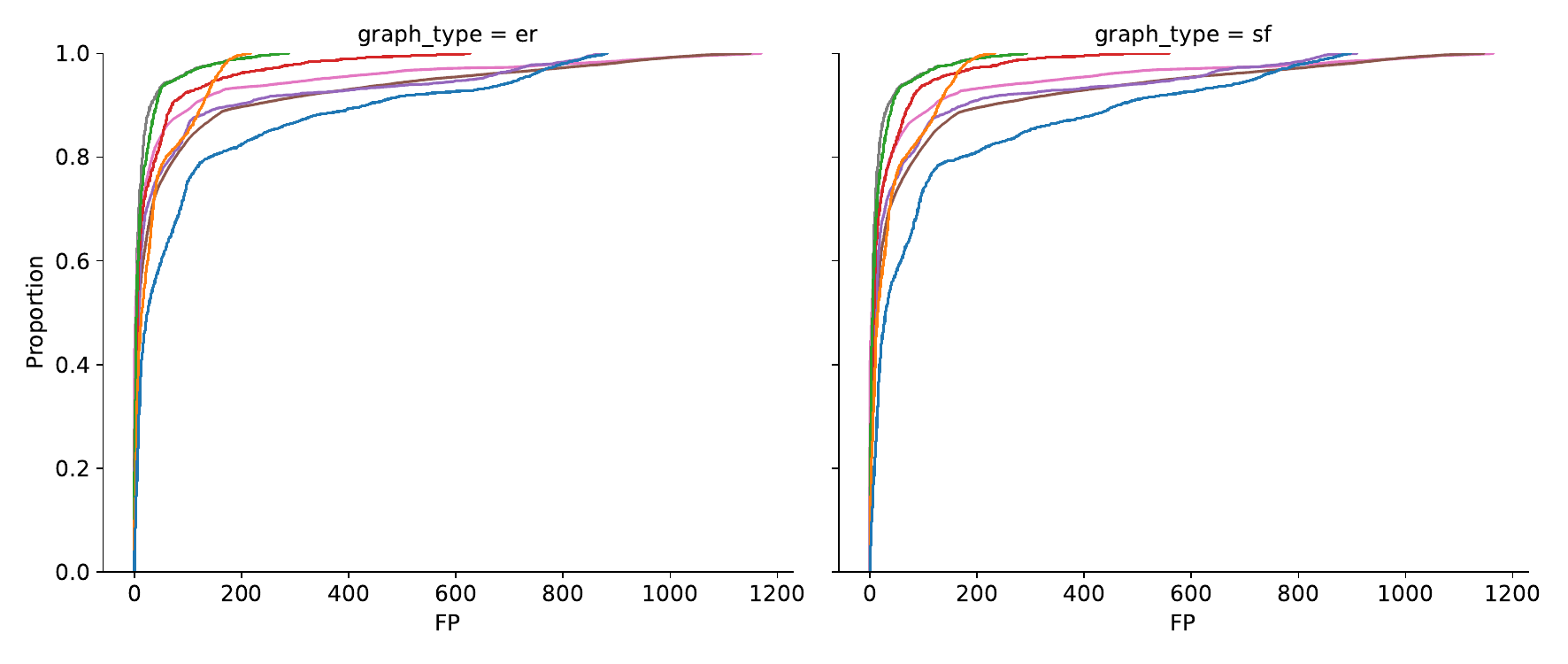}}
    \subfigure[\textit{graph\_p}]{\includegraphics[width=0.7\textwidth]{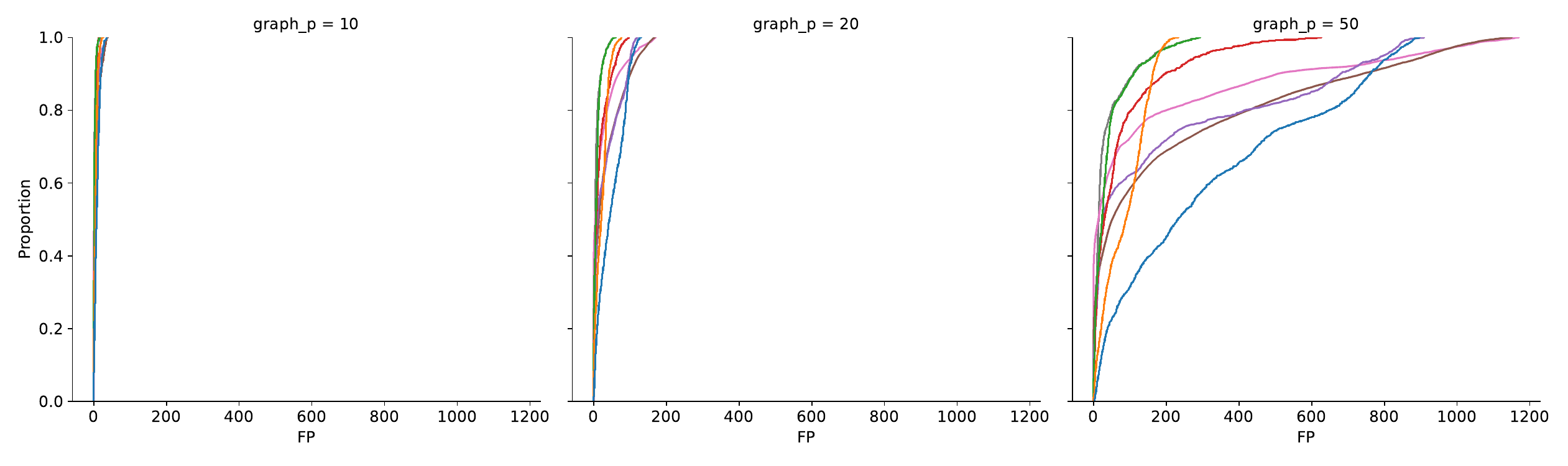}}
    \caption{Distributions of false positive (FP) performances across all hyperparameters, grouped by SEM types (\textit{data\_sem}), sample size (\textit{data\_n}), edge density (\textit{graph\_d}), graph type (ER or SF) and number of nodes (\textit{graph\_p}).}
    \label{fig:dist_fp}
\end{figure}

\begin{figure}[htbp]
    \centering
    \subfigure[\textit{data\_sem}]{\includegraphics[width=0.49\textwidth]{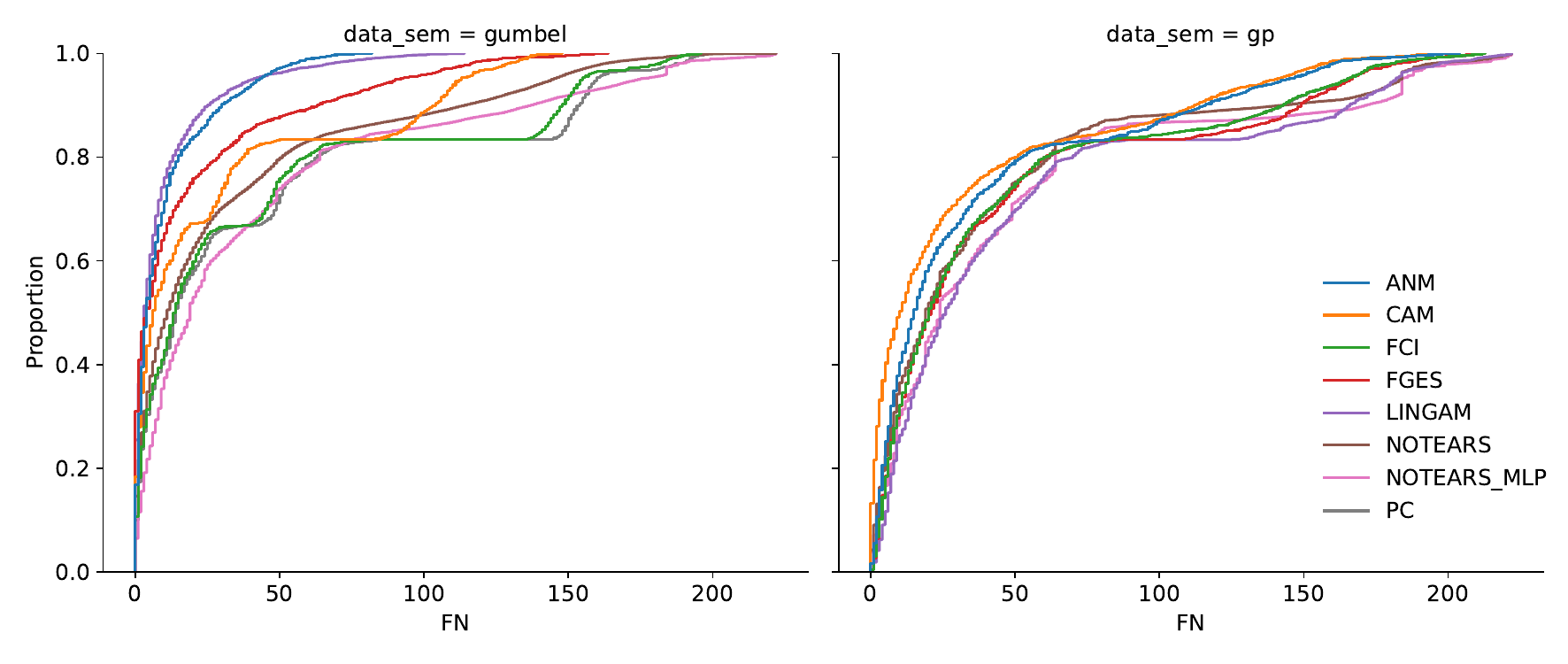}}
    \subfigure[\textit{data\_n}]{\includegraphics[width=0.49\textwidth]{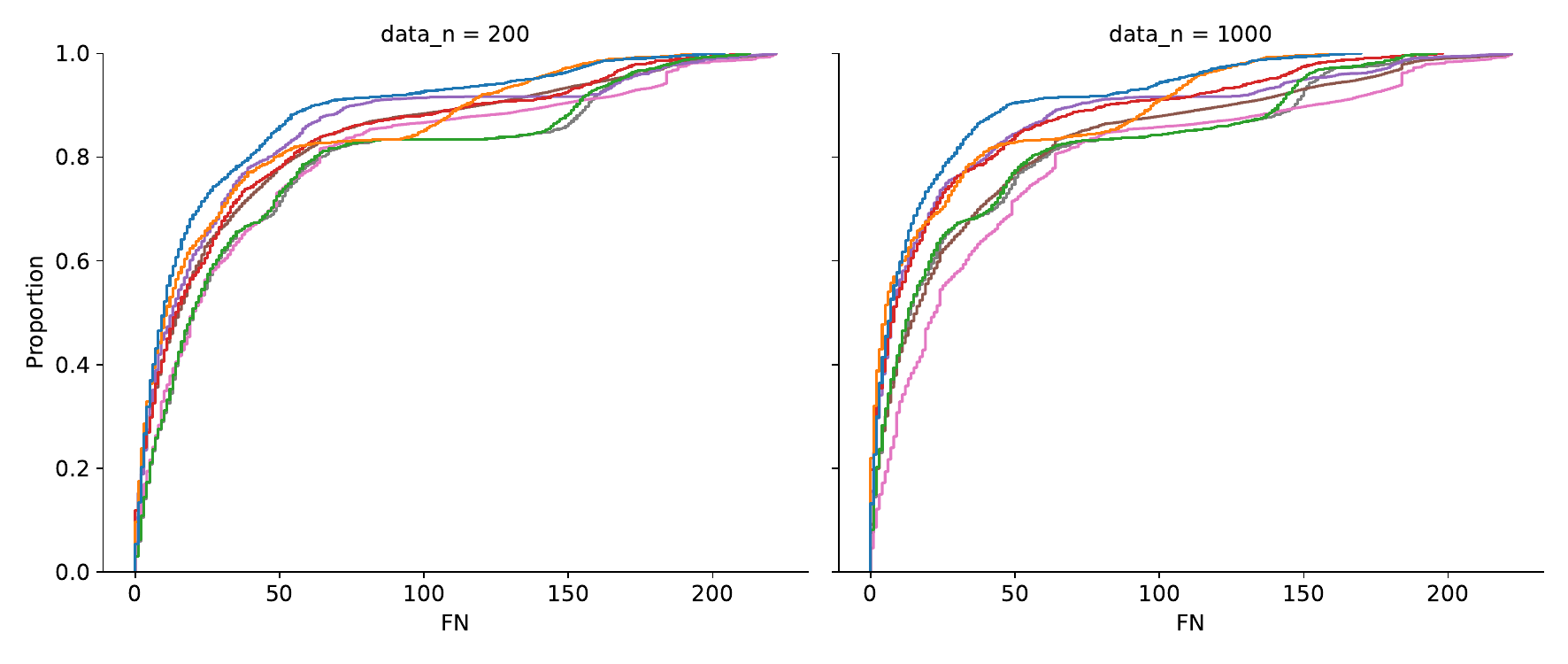}}
    \subfigure[\textit{graph\_d}]{\includegraphics[width=0.49\textwidth]{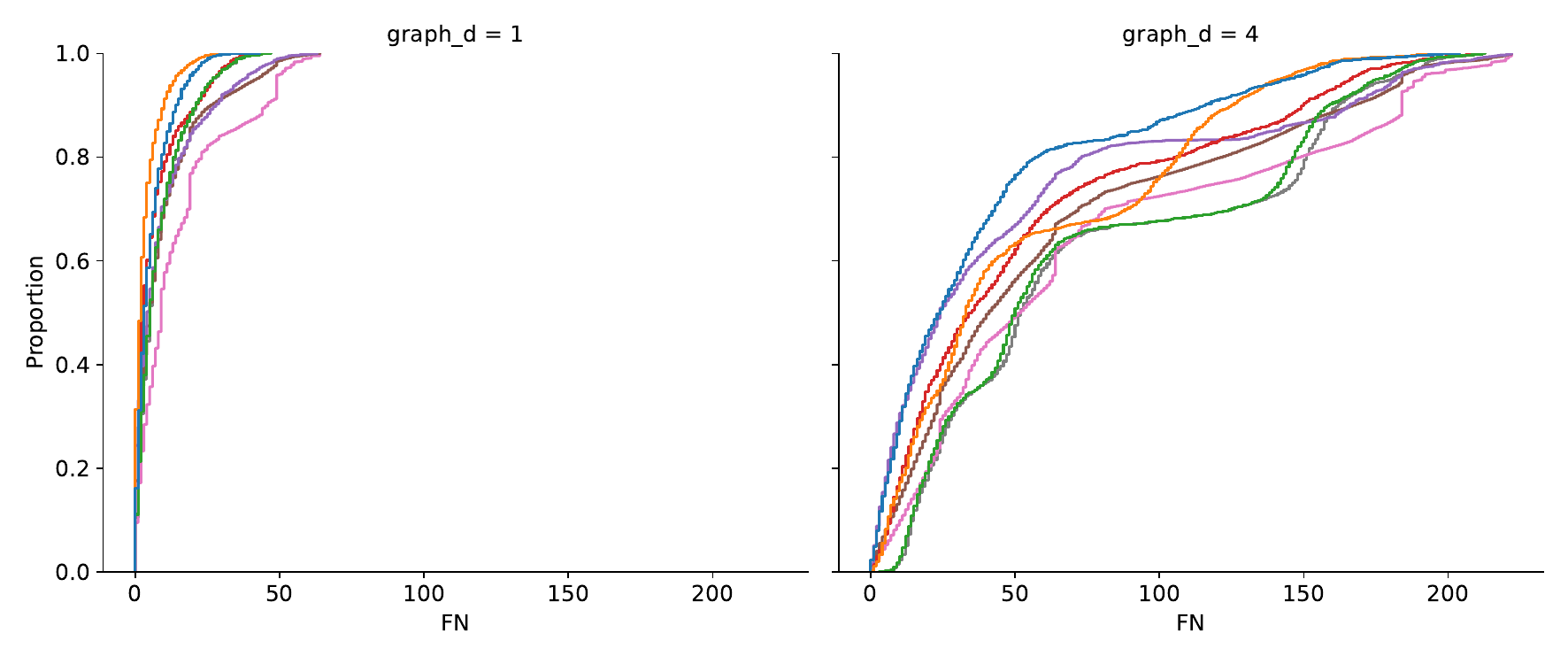}}
    \subfigure[\textit{graph\_type}]{\includegraphics[width=0.49\textwidth]{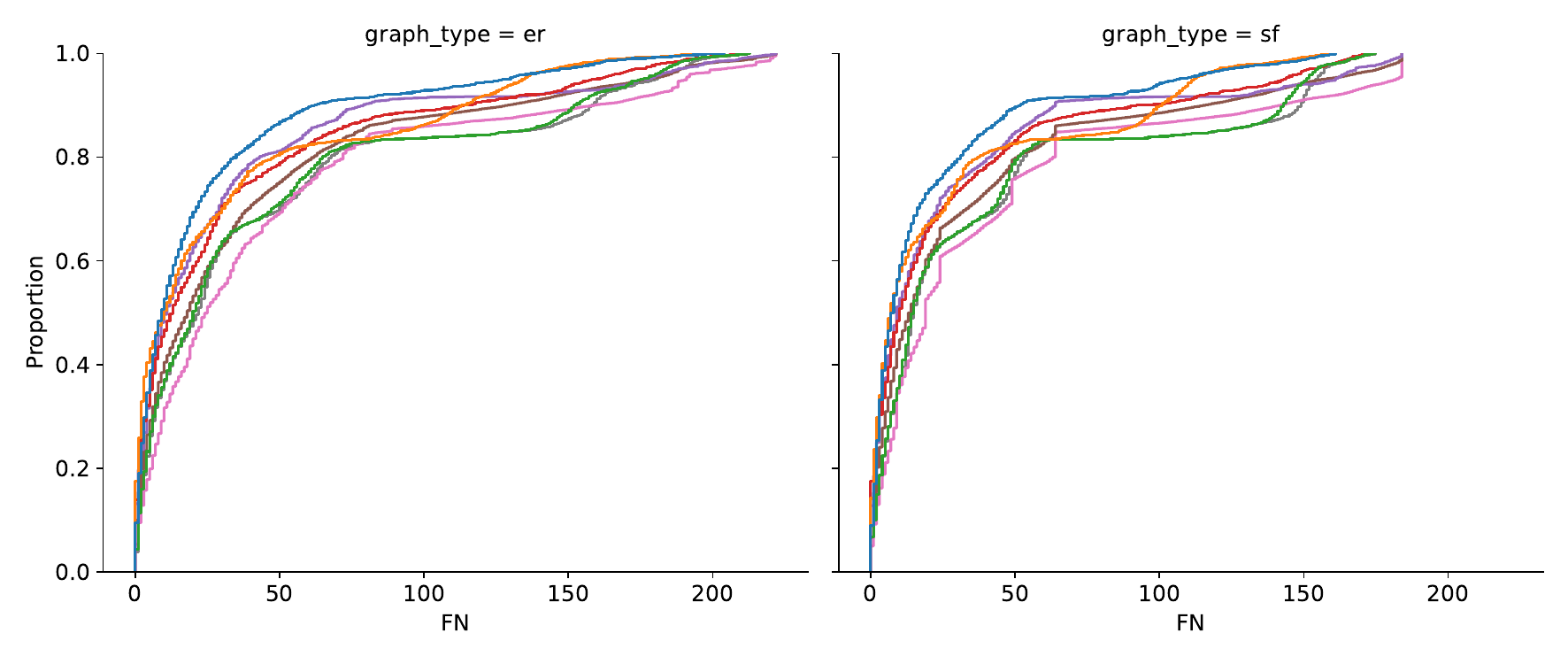}}
    \subfigure[\textit{graph\_p}]{\includegraphics[width=0.7\textwidth]{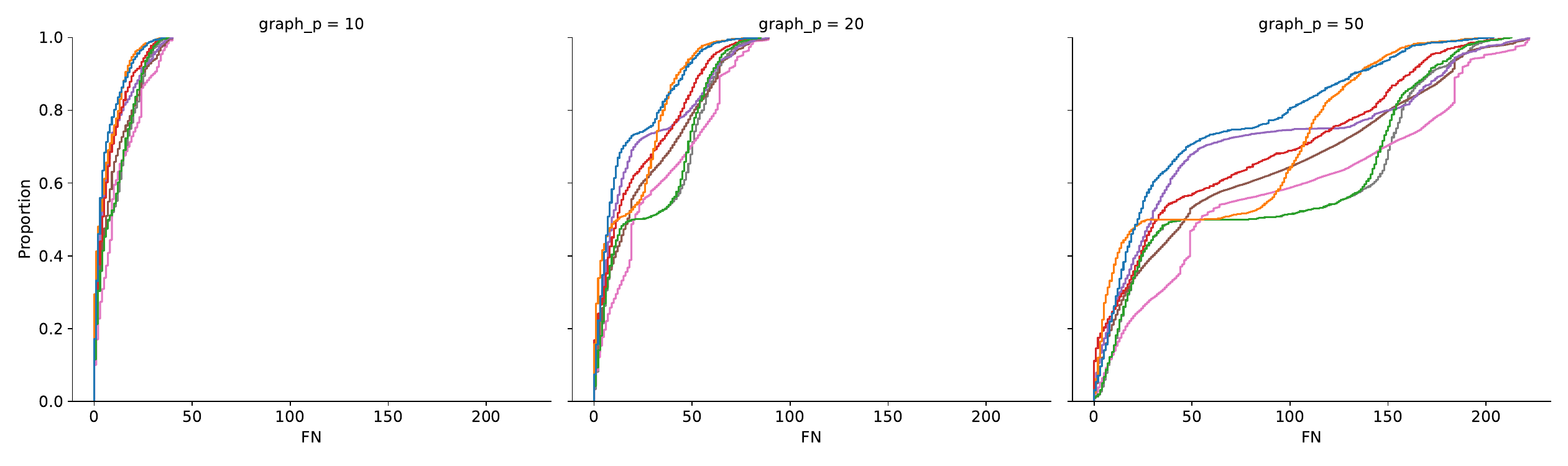}}
    \caption{Distributions of false negative (FN) performances across all hyperparameters, grouped by SEM types (\textit{data\_sem}), sample size (\textit{data\_n}), edge density (\textit{graph\_d}), graph type (ER or SF) and number of nodes (\textit{graph\_p}).}
    \label{fig:dist_fn}
\end{figure}

Figures \ref{fig:dist_shd}, \ref{fig:dist_fp} and \ref{fig:dist_fn} complement Figure \ref{fig:dist_combined} from the main content by showing the results for other types of DGPs.

\newpage
\subsection{Winning Algorithms vs. Simulation Properties}
\begin{figure}[htbp]
    \centering
    \includegraphics[width=\textwidth]{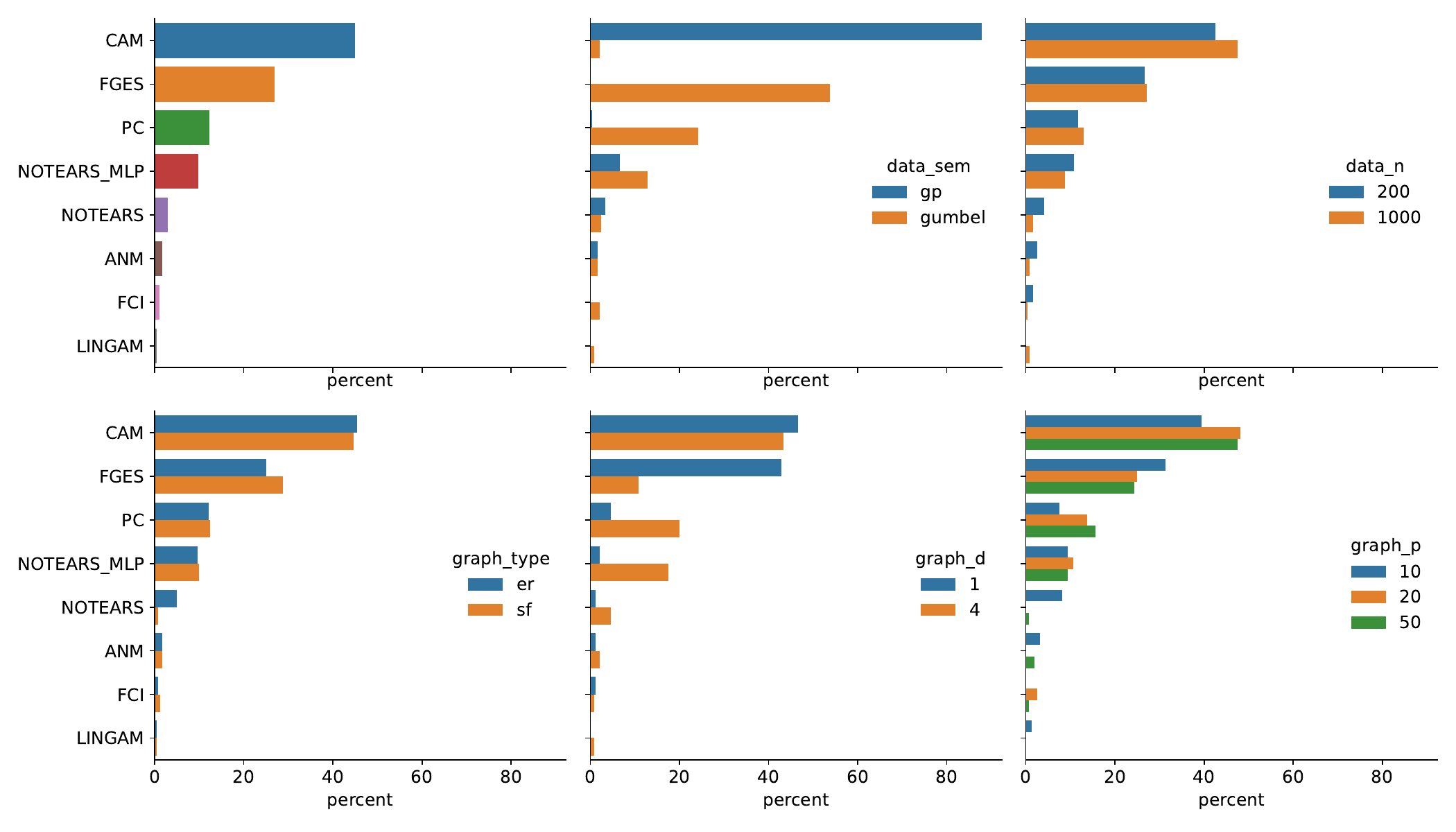}
    \caption{Percentage of wins per algorithm across different DGPs.}
    \label{fig:ms}
\end{figure}

Figure \ref{fig:ms} complements Table \ref{tab:ms_h} from the main content but removes hyperparameters from the picture in order to analyse how DGP properties alone affect winning odds of the algorithms. The results presented here involve the use of the best hyperparameter values. From this perspective, it is clear that no algorithm is the best under all conditions, and that SEM types involved and edge density have the strongest impact on the winning odds.

\subsection{Winning Algorithms vs. Hyperparameter Quality}
\begin{figure}[htbp]
    \centering
    \includegraphics[width=\textwidth]{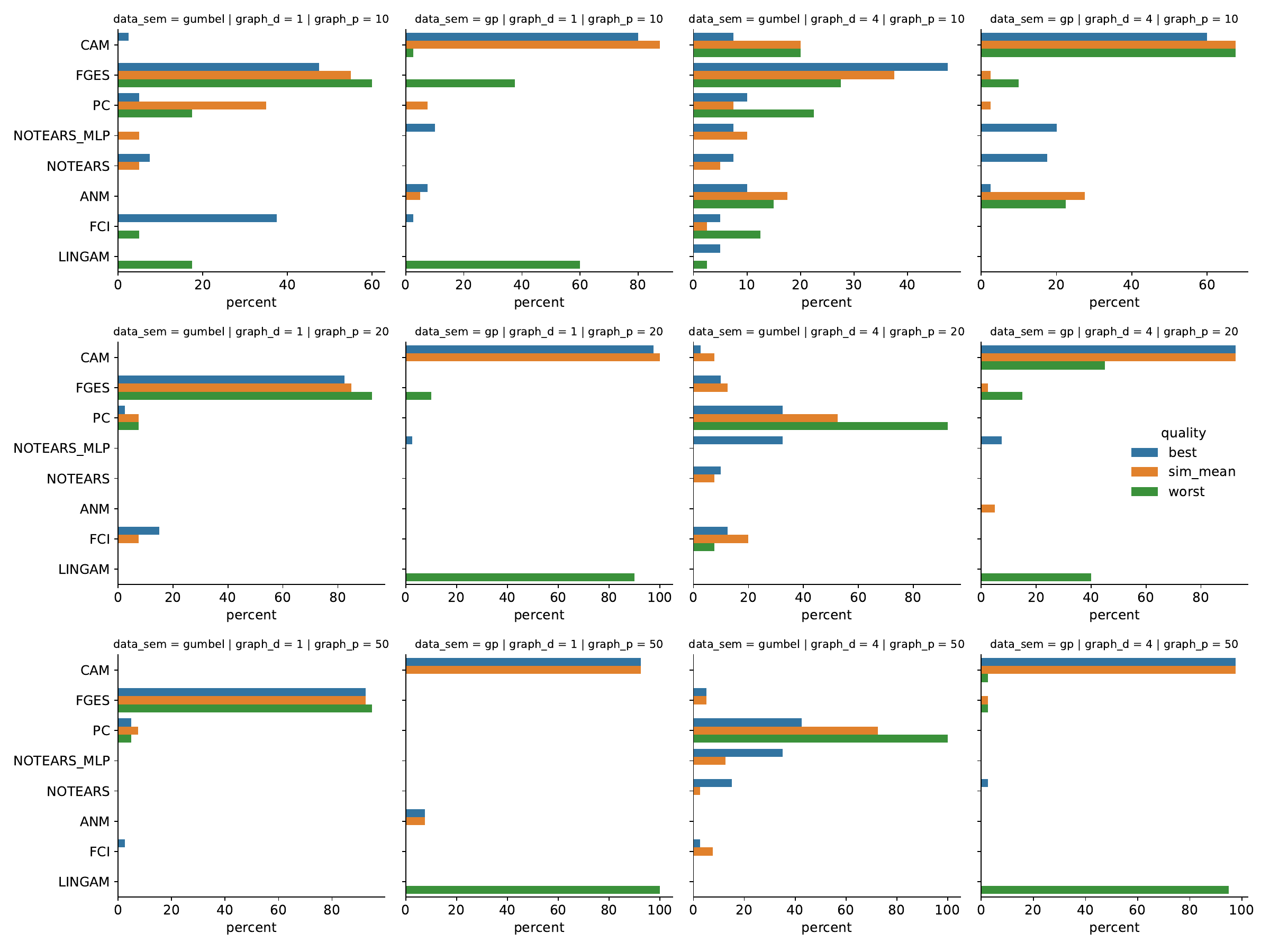}
    \caption{Percentages of winning algorithms under different DGP and hyperparameter conditions.}
    \label{fig:ms_h_combined}
\end{figure}

Figure \ref{fig:ms_h_combined} forms the basis for Table \ref{tab:ms_h} from the main content. It presents winning percentages of algorithms across different DGP types, from which Table \ref{tab:ms_h} was derived.

\newpage
\section{A Guide To Algorithm Selection}\label{app:guide}

\subsection{General Recommendations}
\begin{itemize}
    \item Current algorithms seem to work reasonably well for sparse graphs of up-to 20 nodes. Bigger graphs (50 nodes) are also possible to solve, but more data might be required ($10,000$ samples) to achieve good accuracy. The accuracy of recovered structures drops dramatically for dense graphs.\\ \textbf{Recommendation: Stick to sparse graphs with up-to 20 nodes (moderate amount of data) or up to 50 nodes (a lot of data). Avoid dense graphs.}
    \item No single algorithm is the best option for all problems. Some perform the best under very specific conditions.\\ \textbf{Recommendation: Choose an algorithm that is the most likely to accurately solve the problem at hand based on assumptions derived from data.}
    \item The best choice of an algorithm may depend not only on graph and data properties, but also on the availability of quality hyperparameters. This is because algorithms vary in robustness to misspecified hyperparameters.\\ \textbf{Recommendation: When selecting the best algorithm for the problem at hand, take into account the type of hyperparameters that are available and algorithm's robustness to misspecified hyperparameters.}
\end{itemize}

\subsection{Hyperparameter Selection Strategies}

Optimal hyperparameters are almost never available in structure recovery problems due to inaccessible ground truth. Some methods provide scores that can be used to decide whether a set of hyperparameters is better than others for the same algorithm. However, this strategy cannot be used to compare different algorithms to each other, as they are likely to use different score metrics (while some use none at all). In addition, in order to use those scores, an algorithm's internal code must be modified in most cases, creating a substantial barrier to practitioners.

Thus, default hyperparameters might be a reasonable selection strategy as they often work almost as well as the optimal ones. This is especially the case if the recommended defaults have been derived from data problems similar in nature to the problem at hand.

If, however, the problem to solve is believed to be fairly unique, blindly using default hyperparameters without considering other factors might not be safe. In this case, the best course of action (assuming hyperparameter tuning is not an option) might be to still use default hyperparameters but choose the algorithm that is the most robust to hyperparameter misspecification under specific graph and data conditions that are believed to be applicable to the problem at hand.

\subsection{How to Select Algorithms}

\begin{figure}[hp]
    \centering
    \includegraphics[width=\textwidth]{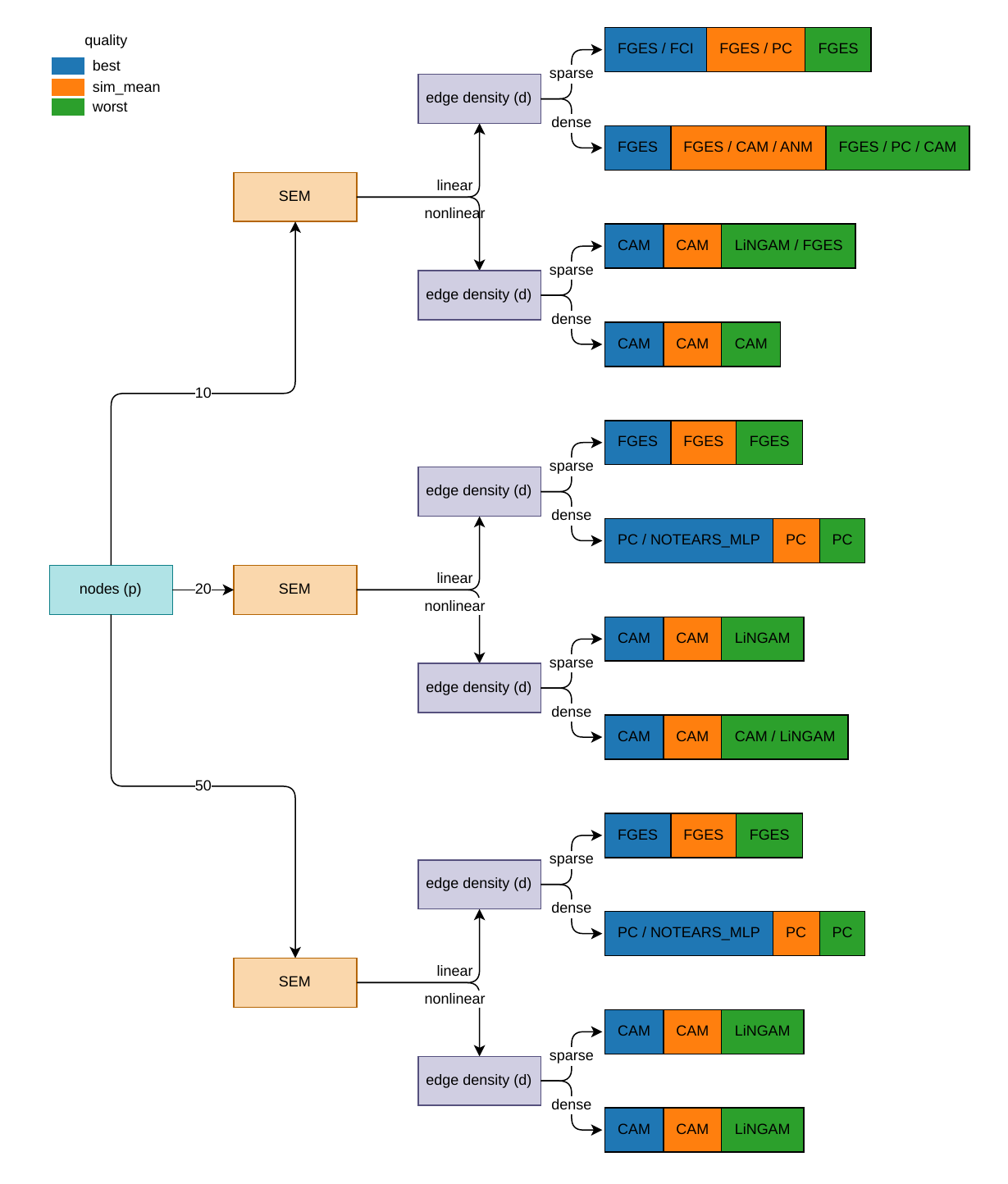}
    \caption{Recommended algorithm choices based on the number of graph nodes, SEM types and edge density in the graph. The final choice depends also on the type of hyperparameters available -- see `quality' colours. In case there is no clear winner, multiple choices are provided in the order of higher winning percentage.}
    \label{fig:alg_choice}
\end{figure}

Figure \ref{fig:alg_choice} summarises the best algorithm choices based on Figure \ref{fig:ms_h_combined}, and which takes into consideration data and graph properties as well as the types of hyperparameters available.

\end{document}